%% file: neurips_main_arxiv.tex
\documentclass{article}

\usepackage[preprint]{neurips_2025}

\input{jam_header} 
\usepackage{algorithm}
\usepackage{algorithmic}

\usepackage[utf8]{inputenc} 
\usepackage[T1]{fontenc}    
\usepackage{hyperref}       
\usepackage{url}            
\usepackage{booktabs}       
\usepackage{amsfonts}       
\usepackage{nicefrac}       
\usepackage{microtype}      
\usepackage{xcolor}         

\usepackage{natbib} 

\title{Higher-Order Causal Structure Learning \\ with Additive Models}

\author{%
  James Enouen$^{1,*}$, Yujia Zheng$^2$, Ignavier Ng$^2$, Yan Liu$^1$, Kun Zhang$^{2,3}$ \\
  $^1$University of Southern California, Los Angeles, CA $\quad$
  $^2$Carnegie Mellon University, Pittsburgh, PA \\
  $^3$Mohamed bin Zayed University of Artificial Intelligence, Abu Dhabi, UAE
}

\usepackage[dvipsnames]{xcolor}

\newcommand\blfootnotetext[1]{%
  \begingroup
  \renewcommand\thefootnote{}\footnotetext{#1}%
  \addtocounter{footnote}{-1}%
  \endgroup
}

\begin{document}
\maketitle

\begin{abstract}
{
    {Causal structure} learning has long been the central task of inferring causal insights from data.
    Despite the abundance of real-world processes exhibiting higher-order mechanisms, however,
    an explicit treatment of interactions in causal discovery {has received little attention}.
    In this work, we focus on extending the causal additive model (CAM) to additive models with higher-order interactions.
    This second level of modularity we introduce to the structure learning problem is most easily represented by a directed acyclic hypergraph which extends the DAG.
    We introduce the necessary definitions and theoretical tools to handle the novel structure we introduce and
    then provide identifiability results for the hyper DAG,
    extending the typical Markov equivalence classes.
    We next provide 
    {insights} 
    into why learning the more complex hypergraph structure  may actually lead to better empirical results.
    In particular, more restrictive assumptions like CAM correspond to easier-to-learn hyper DAGs and better finite sample complexity.
    We finally develop an extension of the greedy CAM algorithm which can handle the more complex hyper DAG search space and demonstrate its empirical usefulness in synthetic experiments. 
}
\end{abstract}

\section{Introduction}\label{sec:intro}
\blfootnotetext{$^*$Correspondence to \texttt{enouen@usc.edu} }

Causal structure learning aims to infer the underlying causal relationships among given variables from observational or interventional data~\citep{spirtes2001causation}, which is crucial for understanding complex systems and has been widely applied in different fields, including biology~\citep{sachs2005causal} and Earth system science~\citep{runge2019inferring}. Various approaches have been developed for causal discovery, including constraint-based, score-based, and functional causal model-based methods~\citep{glymour2019review}.

Constraint-based methods, such as the PC~\citep{spirtes1991pc} and FCI~\citep{spirtes2001causation} algorithms, rely on conditional independence tests to identify the causal structure. Score-based methods, on the other hand, optimize a scoring function, such as the Bayesian Information Criterion (BIC)~\citep{schwarz1978estimating}, to find the best causal structure~\citep{chickering2002optimalGES,singh2005finding,yuan2011learning,bartlett2017integer}. Both constraint-based and score-based approaches can only identify the underlying causal structure up to Markov equivalence~\citep{spirtes2001causation}, indicating that they cannot distinguish between different structures that encode the same set of conditional independence relationships.

Functional causal model-based methods address this limitation by introducing proper functional assumptions on the causal relationships, thus enabling the identification of the whole DAG. Examples include the linear non-Gaussian model~\citep{shimizu2006lingam}, additive noise model (ANM)~\citep{hoyer2008nonlinearCDwithAdditiveNoiseModels}, post-nonlinear causal model~\citep{zhang2009identifiability}, heteroscedastic noise model (HNM)~\citep{xu2022causal,immer2023identifiability}, and causal additive model (CAM)~\citep{buhlmann2014CAM}. Among these, CAM incorporates 
two types of additive models (the additive noise model and the generalized additive model),
assuming that the causal relationships are additive in the variables, which, despite being more restrictive than a general ANM framework, has been shown to achieve superior performance in various empirical studies~\citep{lachapelle2020granDAG,zheng2020learning,ng2022masked,rolland2022SCORE_leafDetectionAlgorithm}, partly owing to its improved statistical power.

In this work, we revisit the additive structural assumption of CAM to also incorporate higher-order interactions,
extending the causal model to explicitly consider higher-order causal mechanisms. 
Higher-order mechanisms are known to exist in a variety of real-world processes \citep{battiston2020networksBeyondPairwiseInteractions},
and are critical for modeling a number of different scientific phenomena
including
dynamical systems \citep{majhi2022dynamicsOnHigherOrderNetworksAReview},
bioscience \citep{taylor2015introductionHOmotivation_HOgeneticInteractions,gaudelet2018introductionHOmotivation_PPIsAndHypergraphlets},
neurobiology \citep{amari2003higherOrderNeurons,petri2014introductionHOmotivation_simplicialFiltrationOnNeruons},
and social networks \citep{freeman1993introductionHOmotivation_socialCliques,deArruda2020introductionHOmotivation_socialContagionHypergraphs}.
Nevertheless, previous approaches to causality fail to adequately represent the higher-order mechanisms which could be at play in real-world data.
Most algorithms take an all-or-nothing approach,
either (a) directly following CAM-like assumptions of no interactions or (b) modeling all possible interactions between the parents of a child node.



Instead, we find that a directed hypergraph can succinctly represent the necessary structure to interpolate between the simplicity of CAM and the complexity of the full ANM.
Specifically, our major contributions are as follows:
\begin{enumerate}
\item We develop the theoretical extension from graphs to hypergraphs across three total settings (undirected graphical models, classical DAG models, and additive noise models), and prove the identifiability of the hypergraph structures we introduce.
\item We develop an algorithm for learning the hyper DAG alongside its structural equations directly from data, extending the greedy algorithm for CAM, and showing improved performance over existing approaches on data specifically containing higher-order variable interactions.
Experiments on simple synthetic data demonstrate that most algorithms struggle to capture higher-order interactions, calling into question the practical learnability of real-world systems with higher-order interactions.
\end{enumerate}





\section{Hypergraph Methods}
\label{sec:definitions}

In this work we will introduce three different generalizations to existing structure learning approaches which extends the existing graphical representations (Markov networks, Bayesian networks, etc.) to their corresponding hypergraphical representations:
\begin{enumerate}
    \item Undirected hypergraphical models
    \item Directed hypergraphical models for discrete variables (classical regime)
    \item Directed hypergraphical models for continuous variables (additive noise model)
\end{enumerate}

We will first introduce the `hyper Markov property' which will be respected by distributions which are `Markov' with respect to a given hypergraph, rather than Markov with respect to a given graph.
We emphasize that since hypergraphs are a strict generalization of existing graphical models, we can see this hyper DAG or HDAG structure as an intermediate level of structure between the DAG and the SEM (structural equation model).
In that sense, we write:
\begin{align}
\text{DAGs} \preccurlyeq \text{HDAGs} \preccurlyeq \text{SEMs}
\end{align}
In what follows,
we will demonstrate that this more fine-grained structure is not only identifiable directly from data, but also that this perspective allows for greater insights into the identifiability of different hypergraphs (and hence graphs) using finite observations rather than the population limit.

We will assume throughout this work that we are in the case of fully observed variables (`causal sufficiency').
Moreover, we will assume for convenience that the density in \ref{subsec:definition_undirected} and \ref{subsec:definition_directed_classical} is strictly positive to ensure (a) that there is no confusion caused by switching between the pairwise, local, and global properties;
and (b) that the score-based definitions we introduce on the log-probability face no ambiguities in regions of zero density.
Extensions are straightforward with appropriate care.



\subsection{Undirected Models}
\label{subsec:definition_undirected}
%
Let us write $X\in\bbR^d$ for some number of dimensions $d\in\bbN$.
We will later choose to restrict to discrete, continuous, or mixed $X$ as appropriate.
We write an \emph{undirected} graph as $\cG'=(V,E')$ and \emph{undirected} hypergraph as $\cH'=(V,H')$, where we take the vertices as $V=[d]:=\{1,\dots,d\}$, the undirected edges as $E'\subseteq \{(i,j) : i\neq j\in V \}$, and the undirected hyperedges as $H' \subseteq \{ S \subseteq V \}$.
We will sometimes abuse notation and write $(i,j)\in\cG'$ to mean $(i,j)\in E'$ and similarly for $\cH'$.
(Note that we are reserving the unprimed versions for the directed versions.)


\begin{definition} \emph{Undirected Markov Property}.
\citep{koller2009probabilisticGraphicalModels}
Let us take $N(i)$ to denote the neighbors of $i\in V$.
We may say that some distribution $p_X(\vec{x})$ is (locally) Markov with respect to some undirected graph $\cG'$ if it holds for any $i$ that  ``$X_i \indep X_{V - N(i) - \{i\}} \oct|\oct X_{N(i)}$'', where -- denotes set minus.
Preparing for our focus on additive models of the log probability, this can equally be required in logarithmic form:
\begin{align}
    \label{eqn:undirected_markov_cond_indep_score_function}
    p_X(\vec{x}) &= p_{N(i)}(\vec{x}_{N(i)}) 
    \hspace{0.2em}\cdot\hspace{0.2em}  p_{i}(x_i | \vec{x}_{N(i)}) 
    \hspace{0.2em}\cdot\hspace{0.2em}  p_{V-N(i)-\{i\}}(\vec{x}_{V-N(i)-\{i\}} | \vec{x}_{N(i)}) 
    \\
    \xi_X(\vec{x}) := \log p_X(\vec{x}) &= \xi_{N(i)}(\vec{x}_{N(i)}) + \xi_{i}(x_i | \vec{x}_{N(i)}) + \xi_{V-N(i)-\{i\}}(\vec{x}_{V-N(i)-\{i\}} | \vec{x}_{N(i)})
\end{align}
where there exists some conditional probabilities $p_i$ and $p_{V-N(i)-\{i\}}$ or some conditional log probabilities $\xi_i$ and $\xi_{V-N(i)-\{i\}}$ such that these equations hold true.
This can be additionally written in terms of the clique representation, when we write all cliques of the graph as $Cl(\cG') = \{ S \subseteq V : S \oct\text{is a clique in}\oct \cG'\}$,
as follows:
\begin{align}
    \label{eqn:undirected_markov_clique_score_function}
    \vspace{-0.3cm}
     \log p_X(\vec{x})  =: \xi_X(\vec{x}) = \sum_{S \in \text{Cl}(\cG')} \xi_S(x_S) 
\end{align}
\end{definition}
\begin{definition} \emph{Undirected Hyper-Markov Property}.
It is now straightforward to generalize this property to hypergraphs as follows:
\begin{align}
    \vspace{-0.3cm}
    \label{eqn:undirected_hyper_markov}
     \log p_X(\vec{x})  =: \xi_X(\vec{x}) = \sum_{S \in \cH'} \xi_S(x_S) 
\end{align}
That is, we write the hypergraph edges as specifically representing the energy terms in the log-probability function.
It is straightforward to verify that this is strictly more general than hypergraphs which can be created as a result of the maximal clique structure of a typical graph.
Nonetheless, in what follows we hope to focus on the identifiability as well as the usefulness of this finer-grained structure for graphical models.
\end{definition}

\subsection{Directed Classical Models}
\label{subsec:definition_directed_classical}
%
%
%
We will write a directed graph as $\cG = (V,E)$ and a directed hypergraph as $\cH = (V,H)$ where the directed edges are $E \subseteq  \{(k,j) : k\neq j\in V \}$ and the directed hyperedges are $H \subseteq \{(S,j) : j\in V, S \subseteq (V-j) \}$.
That is, we are assuming that each hyperedge has only one "out arrow" and up to $|S|$ "in arrows".
It is hoped the purpose for this {restriction} is relatively clear in the context of a causal diagram which must use several parents to generate a single child.
We write the `parents of $j$ in $\cG$' as $\text{Pa}_\cG(j)=\{ k\in[d] : (k,j)\in\cG \}$ and the `hyperparents of $j$ in $\cH$' as $\text{HypPa}_{\cH}(j) = \{ S : (S,j) \in \cH \}$, where the dependence on $\cG$ and $\cH$ will be dropped when obvious.
As can be seen from Equation \ref{eqn:classical_HDAG_Markov} below,
we implicitly assume hierarchy on the hyperedge set via ``$(S,j)\in H, T\subseteq S \implies (T,j)\in H$''.

\begin{definition} \emph{Directed Markov Property}.
\citep{koller2009probabilisticGraphicalModels}
Here, we may once again recall the classical Markov property with respect to a DAG to be written as:
\begin{align}
\label{eqn:classical_DAG_Markov_cond_prob}
    \log p(\vec{x}) = \sum_{i=1}^d \log p(x_i | \vec{x}_{\text{Pa}(i)}) 
    = \sum_{i=1}^d \theta(x_i | \vec{x}_{\text{Pa}})
\end{align}
It is very easy to see that we may rewrite this using extraneous functions as:
\begin{align}
\label{eqn:classical_DAG_Markov_Z_score}
    \log p(\vec{x}) = \sum_{i=1}^d \Big( \theta(x_i ; \vec{x}_{\text{Pa}(i)}) - \cZ(\vec{x}_{\text{Pa}(i)}) \Big) 
\end{align}
where it is now the case that we do not have the $\theta$ energy terms explicitly representing a conditional distribution, but are instead arbitrary functions which are then set to the proper normalization via the $\cZ$ function.
It can be seen that the $\cZ$ function does not explicitly depend on the value of $x_i$, but normalizes to a distribution based on only the parents alone.
The extraneous $\theta$ functions which are written as all subsets are useful for the next step generalizing to hypergraph structures.
\end{definition}

\begin{definition} \emph{Directed Hyper-Markov Property}.
We follow the structure above from the typical DAG framework, but replace the fully-connected parent structure with the more nuanced hypergraph structure.
In particular, the energy terms in each of the conditional distributions are replaced with a more specific additive model structure, rather than assuming there is a generic function:
\begin{align}
\label{eqn:classical_HDAG_Markov}
    \log p(\vec{x}) = \sum_{i=1}^d\Big(  \Big( \sum_{S\in \text{HypPa}(i)} \theta(x_i ;\vec{x}_{S})\Big)  - \cZ(\vec{x}_{\text{Pa}(i)}) \Big)  
\end{align}
It is straightforward to see that this again strictly generalizes the structures which are representable by the typical DAG.
Importantly, this includes structures which are not easily represented through adding latent variables to existing graphical approaches.
It can also be seen from equation \ref{eqn:classical_HDAG_Markov} that we may restrict our attention to hypergraphs which are \emph{hierarchical},
meaning $S \in \text{HypPa}_{\cH}(j)$ and $T \subseteq S$ implies $T \in \text{HypPa}_{\cH}(j)$ (much like a simplicial complex).

In particular, taking the hyperparents to be all subsets of the parents recovers the previous functional form (Figure \ref{fig:hypergraph_teaser}f).
However, other structures mimicking CAM and LiNGAM type assumptions are also possible (Figure \ref{fig:hypergraph_teaser}d).
Further HDAGs beyond these two existing in the literature are also possible (Figure \ref{fig:hypergraph_teaser}e).
%
We will further assume `causal minimality' of the HDAG meaning the hyperparent set is downwards closed w.r.t subsets and each maximal element has a nontrivial $\theta$ function.
See the discussion and proofs in the appendix for further details.
\end{definition}

It is also relatively clear to see how this \emph{directed} hyper-Markov property overlaps with the \emph{undirected} hyper-Markov property, perhaps moreso than the typical Markov formulation.
Moreover, it becomes clear that the moralized graph corresponds to including the $\cZ$ terms whereas the skeleton corresponds to including only the $\theta$ terms, see also Table \ref{tab:notation_table}.
We will make this connection more clear in  {Section} \ref{subsec:identifiability_directed_classical},
where we show identifiability of the HDAG up to its hyper Markov equivalence class (HMEC).


\subsection{Directed Additive Noise Models}
\label{subsec:definition_directed_ANM}
For continuous variables, we will generate data from the additive noise model (ANM),
meaning that all variables are a deterministic function of their parent variables, plus an additive noise term.

\begin{definition}
\emph{Additive Noise Model}. 
\citep{peters2014continuousANM}
This may be written as:
\begin{align}
    \label{eqn:ANM}
    X_j = f_{\text{Pa}(j)\rightarrow j} (X_{\text{Pa}(j)}) + \eps_j    
\end{align}
Each of the $\eps_j$ are taken to be independent, mean-zero random variables.
\end{definition}


\begin{definition}
\emph{Higher-Order Additive Noise Model}. 
We may once again generalize to the higher-order additive model through the use of the structure encoded by the directed hypergraph.
\begin{align}
    \label{eqn:HO_ANM}
    X_j = \Big( \sum_{S\in\text{HypPa}(j)} f_{S\rightarrow j}\big(x_{S}\big) \Big) + \eps_j    
\end{align}

\end{definition}

\begin{figure*}[!htb]
  \centering
  \includegraphics[width=.97\linewidth]{ {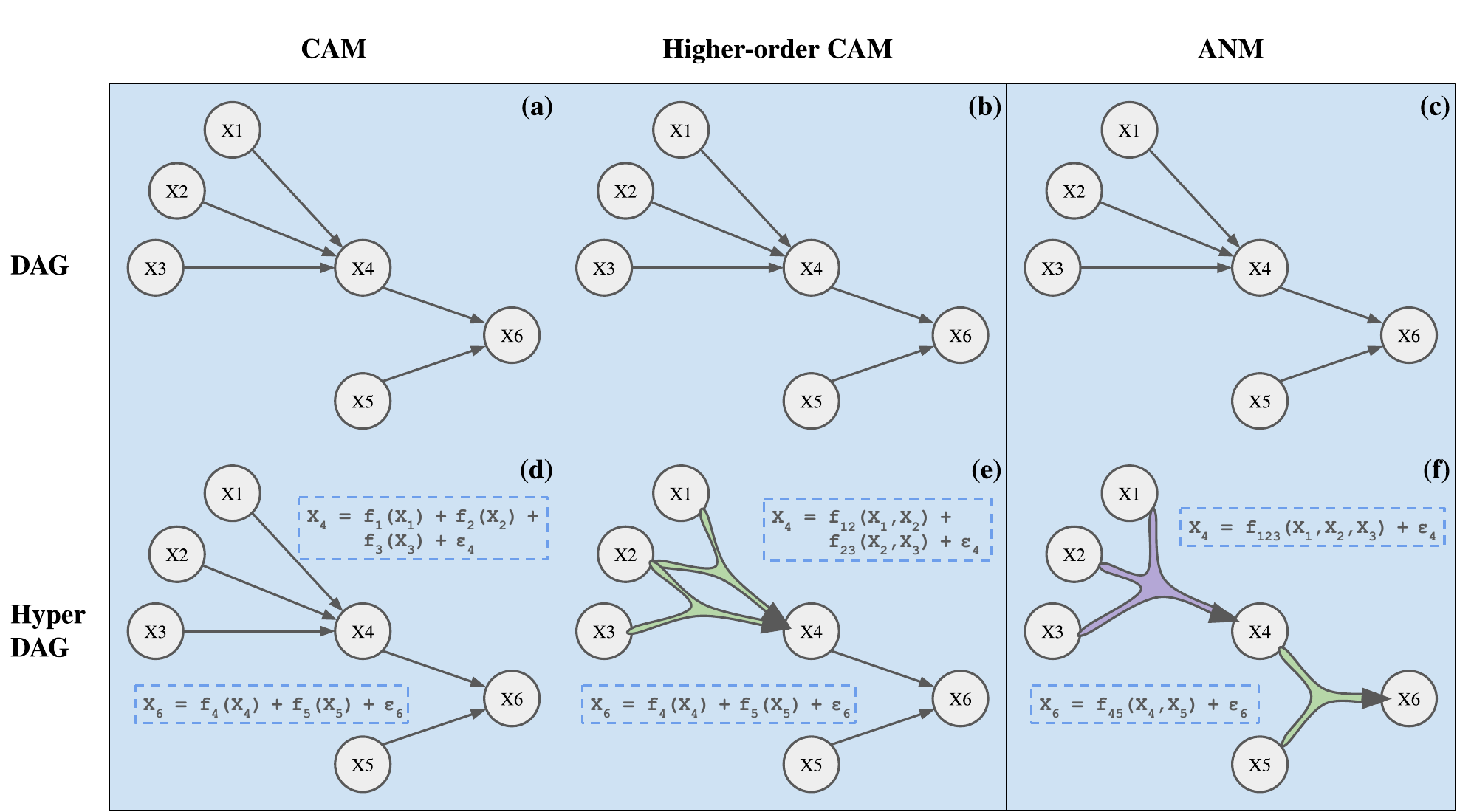} }
  \caption{A depiction of the distinguishing power for hypergraphs corresponding to the same DAG.  The CAM assumption implies moving from panel a to d.  The ANM assumption implies moving from panel c to f.  We propose the more HDAG which includes both CAM and ANM as special cases.
  }\label{fig:hypergraph_teaser}
\end{figure*}

Specifically, we endow the generating function $f_{\text{Pa}(j)\rightarrow j}$ with an additive model structure obeying the hyperparents of the HDAG.
Models like CAM or LinGAM then correspond to using singleton hyperparents, whereas the most general ANM corresponds to using the entire block of parents as the largest hyperparent, as depicted in Figure \ref{fig:hypergraph_teaser}f.
We will follow CAM \citep{buhlmann2014CAM} in assuming Gaussian noise for algorithmic purposes via the minimization of mean-squared error corresponding to the maximization of log-likelihood; however, surprisingly,
we show in Theorem \ref{thm:thm3} that the our settings 2 and 3 actually overlap in the case of additive Gaussian noise.

\section{Structure Identifiability}
\label{sec:identifiability}

\subsection{Undirected Models}
\label{subsec:identifiability_undirected}


First, recall that under our assumptions of fully observed variables and strictly positive density,
the density function is identifiable directly from the observed distribution (under mild assumptions like continuity for the continuous variable case) \citep{rosenblatt1956densityFunctionEstimation}.

Importantly, then, one may only be concerned in measuring the hypergraph structure as described in Section \ref{subsec:definition_undirected}; however, this proves to be equally straightforward.
For the case of graphical models and mixed-type variables, \citet{zheng2023generalizedPrecisionMatrix} write the generalized precision matrix as:
\begin{align}
    \Omega_{i,j} := 
    \Big\| \frac{\partial^2 }{\partial_i \partial_j} \log p(x) \Big\| := 
    \Big( \bbE\Big[ \Big| \frac{\partial^2 \log p(x)}{\partial_i \partial_j}   \Big|^2 \Big] \Big)^{\frac{1}{2}}
\end{align}

In the case of hypergraphical models and discrete variables, \citet{enouen2024completeDecompositionfKLerror} similarly write the existence of `higher-order information' for some subset $T\subseteq[d]$ (where $T\supsetneq \{i,j\}$ is chosen to {imply higher-order}) if it is the case that:
\begin{align}
    \Omega_T := \Big\| \sum_{S\supseteq T} \theta_S(x_S) \Big\| > 0
    \quad\quad\text{where}\quad\quad
    \log p(x) = \sum_S \theta_S(x_S)
\end{align}
A straightforward combination of these approaches are sufficient for recovery of the hyper Markov network or undirected hypergraph.
Nonetheless, our experiments will instead focus on identification of the directed structure as in the following two sections.
Thus, for our purposes it is sufficient to say that the density and log density functions are identifiable directly from the observed distribution.


\subsection{Directed Classical Models}
\label{subsec:identifiability_directed_classical}
Before our main theorem of identifiability extending the result of \citet{vermaPearl1990skeletonsAndColliders},
we must first 
introduce the notion of multi-dependence to extend the typical notion of conditional dependence which is the workhorse of causal structure learning.
We will focus on discrete and finite variables as in the classical case~\citep{verma1993causalModelTechnicalReport,pearl2009causalityBook}; however, most results clearly extend to continuous or mixed variables under mild conditions, and we later discuss one such special case in Theorem \ref{thm:thm3}.

\begin{definition}
\emph{Conditional Multi-{in}dependence}. 
We write that $X_i$ and $X_j$ are dependent if the distribution $\log p(x_i,x_j)$ must be written with a 2D energy term, $\theta_{ij}(x_i,x_j)$, rather than the sum of two 1D energy terms, $\theta_i(x_i) + \theta_j(x_j)$, (corresponding to the product when the log is removed).
We will write that $X_i$, $X_j$, and $X_k$ are tri-dependent (or generally multidependent), if the distribution $\log p(x_i,x_j,x_k)$ must be written with a 3D energy term, rather than the sum of three 2D energy terms,
$\theta_{ij}(x_i,x_j) + \theta_{ik}(x_i,x_k) + \theta_{jk}(x_j,x_k)$.
It can be seen that this does not have a convenient product formulation like the classical case of dependence and independence because of the "mixing" or "torsion" between the three 2D terms.
Nonetheless,
we will attempt to prove the usefulness of such a definition in the following theorem.
Generalization to conditional tests is straightforward.
{In the same way that we write that two graphs are Markov equivalent if they have the same conditional independences, we will write that two hypergraphs are hyper Markov equivalent if they have the same conditional multi-independences.}
\end{definition}

\begin{theorem}
\label{thm:hyper_MECs}
    The HDAG is identifiable up to the hyper Markov Equivalence classes (HMECs), consisting of all HDAGs with the same ``body"  and the same (unshielded) ``multi-colliders", paralleling the existing result identifying DAGs up to their skeleton and (unshielded) colliders.
\end{theorem}

\begin{wraptable}{r}{.638\linewidth}
\begin{minipage}{.95\linewidth}
    \centering
    \caption{Notation for hypergraphs. Details in App. \ref{app_sec:full_definitions}.}
    \label{tab:notation_table}
\resizebox{.95\columnwidth}{!}{
    \begin{tabular}{|c|c|}
    \hline
        DAG terms & HDAG terms \\
    \hline
         $\cG'$, undirected graph & $\cH'$, undirected hypergraph  \\
         $\cG$, directed acyclic graph (DAG) & $\cH$, hyper DAG or HDAG \\
         moralized graph of a DAG & immoralized hypergraph of an HDAG \\
         skeleton of a DAG & body of an HDAG \\
         (unshielded) collider & (unshielded) multicollider  \\
         conditional independence test & conditional multi-independence test  \\
    \hline        
    \end{tabular}
}
\end{minipage}\hfill
\end{wraptable}

\begin{figure*}[!htb]
  \centering
  \includegraphics[width=.97\linewidth]{ {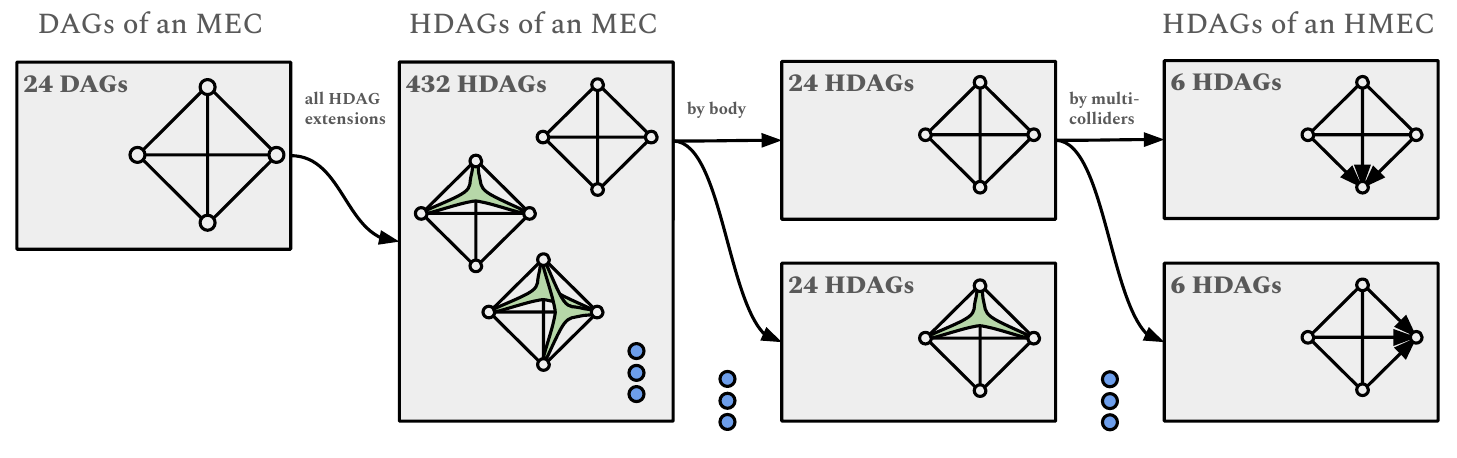} }
  \caption{A gradual refinement of the DAGs within a MEC to a stronger refinement of HDAGs based on Theorem \ref{thm:hyper_MECs} to an HMEC.
  There are $d=4$ variables with a fully-connected DAG structure. 
  The 24 DAGs correspond with the 4! orderings of the four variables.
  The green triangles represents third-degree hyperedges in the body of an HDAG.
  Lack of arrows indicate uncertainty about the orientation of the edges or hyperedges (between members of the same equivalence class).
  }
  \label{fig:dag_to_hdag_mec_to_hmec}
\end{figure*}

In the same sense that a conditional independence test can never eliminate a causal arrow between two variables,
a conditional multi-independence test can never separate a higher-order causal relationship between a set of three or more variables.
Removing the arrowheads from the DAG returns the DAG's skeleton; similarly, removing the arrowheads from the HDAG returns the HDAG's body, see Table \ref{tab:notation_table} and Figure \ref{fig:dag_to_hdag_mec_to_hmec}.
In some sense, this half of the theorem about the ``body identifiability'' immediately states that the structure we introduced is identifiable.

For multicolliders,
recall that a collider occurs when there is a conditional dependence which is broken after marginalizing out the child, or equally a conditional independence which is broken when conditioning on the child.
The multicollider of an HDAG will occur similarly via a multidependence which is broken after marginalizing out the child.
Although collisions between two parents will already be covered, there are cases of three or more parents which are unshielded and can hence be identified from Theorem \ref{thm:hyper_MECs}.
In particular, there are cases which are not identified in the classical setting, see the RHS of Figure \ref{fig:dag_to_hdag_mec_to_hmec}.
This seeming anomaly is in part due to the historical conflation over time between what structure is recoverable from the conditional independence tests vs. what structure is recoverable from the observed distribution.
Indeed, the MEC only describes what is distinguishable via the conditional independence conditions, making it unable to detect what can be seen via the conditional multi-independence test we introduce.

\nocite{chickering1995equivalentBayesianNets}
\nocite{andersson1997characterizationOfMECs}

Previous attempts to incorporate hypergraphs into causality in the classical setting like chain graphs \citep{javidian2020hypergraphPGM} and graphs with latent variables \citep{evans2016hypergraphsForMarginsOfBayesNets}
have failed to fully characterize Markov equivalence.
Another key consequence of this different perspective is the statistical insight.
In particular,
for a $K$-dimensional energy term in the body of an HDAG,
we know that it requires on the order of $\cO(n^K)$ samples to be appropriately learned.
Consequently, without access to infinite samples,
this places further restrictions on the HMEC classes (and hence MEC classes) of `distinguishability under finite samples',
whereas MECs have historically only focused on `distinguishability under infinite samples' as in the asymptotic regime.

\subsection{Directed Additive Noise Models}
\label{subsec:identifiability_directed_ANM}
In this section, we establish identifiability results for recovering the hyper-DAG in the ANM case.
%
For clearer exposition, we first reproduce the arguments of \cite{hoyer2008nonlinearCDwithAdditiveNoiseModels} which shows that, in general position, the additive noise model (ANM) is identifiable.  
We extend their result to a multi-dimensional result which handles the case of multiple parents rather than only the case of one parent node and one child node (slightly different from the extension in Theorem 28 of \cite{peters2014continuousANM}
because it will more easily generalize to the hypergraph result).

\begin{theorem}
    \label{thm:thm1}
    Let the joint probability densities of $\Vec{x}$ and $y$ be given by
    \begin{align}
        p(x_1,\dots,x_d,y) = p_n(y - f(\Vec{x})) \cdot p_x(x_1,\dots,x_d)
    \end{align}
    for some noise density $p_n$, arbitrary density $p_x$, and function $f$.
    Assume further that all such functions are thrice continuously differentiable.
    If there is also a backwards model which treats $x_i$ as the child, 
    \begin{align}
        p(x_1,\dots,x_d,y) = p_{\Tilde{n}}(x_i - g(x_{-i},y)) \cdot p_{x_{-i},y}(x_1,\dots,x_{i-1},x_{i+1},\dots,x_d,y) 
    \end{align}
    for some alternate noise density $p_{\Tilde{n}}$,  arbitrary density $ p_{x_{-i},y}$, and function $g$,
    then it must be case that for $\nu := \log p_n$ and $\xi := \log p_x$, the following differential equation is obeyed
    \begin{align}
        \xi''' = \xi'' \cdot \Big(-\frac{\nu''' f'}{\nu''} + \frac{f''}{f'} \Big) + 
        \Big(-2\nu'' f'' f' + \nu' f''' + \frac{\nu' \nu''' f' f''}{\nu''} - \frac{\nu' f'' f''}{f'}    \Big) 
    \end{align}
    where we use $\xi',\nu',f'$ as shorthand for $\frac{\partial}{\partial x_i} \xi(x_1,\dots,x_d), \quad \nu'(y-f(\vec{x})), \quad \frac{\partial}{\partial x_i} f(x_1,\dots,x_d)$.
\end{theorem}

\begin{corollary}
    \label{thm:cor1}
    Assume further that $\nu'''$ = $0$ and $\frac{\partial^3}{\partial x_i^3} \xi$ = $0$ for all $i\in[d]$.
    If a backwards model exists for a parent $x_i$, then $f$ is linear in the argument $x_i$.
    Further, if a backwards model exists in all parents $x_i$, then $f$ is a multilinear function.
\end{corollary}

\begin{remark}
    First, this is a general position argument which says that in order to be reversible, 
    the SEM must obey these particular constraints which usually do not occur.
    Although this means for an `arbitrary' SEM model we have identifiability, this does not rule out well-known cases including linear+Gaussian where the existence of an equivalent backwards model is unavoidable.
\end{remark}

\begin{remark}
    Second, this result is applied locally to a single child node, rather than the global SCM structure.
    Previous work has partially explored the global structure in the population limit, see \cite{chickering2002optimalGES} and \cite{peters2014continuousANM};
    however, a priori, the constrained solutions space may grow exponentially large, leading to practical limitations in the real-world setting with finite samples.
\end{remark}



We next provide the relevant extension to recover the hypergraphical structure as well.
Indeed, if the DAG structure is identifiable (at least locally), then the hyper DAG structure is also identifiable (at least locally).
This also implies that when the entire DAG structure is identifiable in the ANM case, the entire hyper DAG structure is also identifiable.

\begin{theorem}
    \label{thm:thm2}
    Suppose we have two forward models given by two alternate collections $\cI \subseteq \cP([d])$ and $\cJ\subseteq \cP([d])$, where $\cP$ denotes the power set, with models:
    \begin{align}
        p(y | x_1,\dots,x_d) = p_n(y - \sum_{S\in\cI} f_S(x_S)) & & & &
        p(y | x_1,\dots,x_d) =  p_{\Tilde{n}}(y - \sum_{T\in\cJ} g_T(x_T))
    \end{align}
    Assume that in addition to the assumptions of Theorem \ref{thm:thm1}, we also have that the functions $f_S$ and $g_T$ are differentiable up to order $\max\{ \max_{S\in\cI}\{|S|\}, \max_{T\in\cJ}\{|T|\} \}$, or more simply up to order $d$.
    It then follows that $\cI = \cJ$ and thus the hypergraph structures of the two models are exactly the same.
    Moreover, outside of trivial modifications to the $f$'s and $g$'s, the two functional generating models are exactly the same.
    {The proof technique is the same as Theorem \ref{thm:thm1}.}
\end{theorem}

Finally,
we restrict further to the case of Gaussian noise variables in the ANM to directly recover a global hypergraphical result,
further relating Sections \ref{subsec:definition_directed_classical} and \ref{subsec:definition_directed_ANM}.
\begin{theorem}
    \label{thm:thm3}
    Suppose that all additive noise variables are drawn from a Gaussian distribution, implying that $\nu(\eps) = \frac{-1}{2\sigma^2}\eps^2$ (or some general quadratic form) for all $i\in[d]$.
    Suppose also that we generate data according to the directed hypergraph $\cH$.
    The {undirected, immoralized version} of this hypergraph $\cH'$ is identifiable directly from the data in the sense that $\xi(x_1,\dots,x_d) = \sum_{S\in\cH'} \xi_S(x_S)$ for some arbitrary functions $\xi_S$.
    In other words, the undirected hypergraph from setting 1 is directly identifiable from the distribution and thus settings 2 and 3 overlap for ANMs with Gaussian noise.
\end{theorem}

All full proofs may be found in the appendix.

\color{black}

\section{Algorithm}

Our method for hypergraph discovery and structural equation modeling heavily entwines the schema of CAM \citep{buhlmann2014CAM} and the higher-order interaction techniques of SIAN \citep{enouen2022sian}.
Accordingly, we review both algorithms in much greater detail in Appendix \ref{app_sec:algorithm_details}.
Here in the main body of the paper,
we briefly review the three step procedure introduced by CAM and discuss our HCAM extension to their original approach.
The first stage is a preselection phase which searches for good candidate parents for each potential child node.
The second stage is the bulk of the algorithm, greedily constructing the DAG via including each directed edge one at a time based on the improvement to the log-likelihood.
The third stage is a final pruning stage which does not change the topological order, but simply removes parents which are no longer thought to be relevant.

\subsection{Step 1: Candidate Search}

The first stage of CAM \citep{buhlmann2014CAM} finds candidate edges/ parents by training a GAM regression on each of the $d$ variables.
In their work, this was mainly necessary for them in the high-dimensional setting, where explicit consideration of so $\cO(d^2)$ edges when $d$ is large poses a challenge.
However, for HCAM, 
this stage becomes absolutely critical.
That is because we not only need to look at all $(d^2-d)$ candidate directed edges of CAM, but also all the $\frac{1}{2}(d^3-3 d^2+2d)$ candidate directed tri-edges, as well as higher-order hyperedges, etc.
To consider all hyperedges directly without any heuristic would require considering an exponential number of hyperedges.

Accordingly, 
we first use a deep neural network to `search' for good hyperedges, 
following steps 1 and 2 of SIAN \citep{enouen2022sian}.
That is, for each of the $d$ variables, we regress an MLP DNN to minimize the mean-squared error (Gaussian likelihood).
We then use an XAI technique, called Archipelago \citep{tsang2020archipelago}, to give an importance score for each of the feature interactions involving the other variables.

\subsection{Step 2: Greedy Selection}

The next phase consists of the bulk of the algorithm and can also be considered the most important part and yet most simplistic part.
A greedy heuristic is taken to gradually build the DAG from the initialized empty set of vertices.
For each of the possible edges (or potentially the subset selected in step 1), 
a new likelihood model is trained to simulate adding the edge to the DAG.
Importantly, because of the independence of the ANM noise, this is simply measured as the drop in MSE with and without the additive term.
Gradually, edges are added until some stopping point and the final phase of pruning begins.

In our case, several small adjustments must be made to deal with the case of HDAGs.
Importantly, unlike CAM, HCAM must keep track of both the HDAG and the induced partial-order matrix simultaneously.
Generically, we follow the exact same procedure, training higher-order additive models with each candidate hyperedge to see the improvement in MSE.
We start with $10$ candidates for each of the $d$ variables, based on the ranking provided in step 1, 
and we replenish the candidates if there are ever less than $5$ viable hyperedges per a variable.
This cutdown is able to reduce the number of higher-order SIAN additive models we train, 
which helps in improving the overall runtime.

\subsection{Step 3: Final Pruning}
Finally, the full model is trained end-to-end once more with all of the included edges from step 2.
The final stage simply removes edges corresponding to additive terms which are too close to zero, and thus likely to be useless in the causal model.
Our extensions prunes in the exact same way, with higher-order terms corresponding to higher-order edges, but there is no practical difficulty in doing this extension.
The major difficulty of this part is choosing a threshold which corresponds with a nuisance parameter.
The original CAM work uses a threshold of $0.001$ for the p-values provided alongside the GAM models.
We use neural networks which do not provide p-values and thus somewhat similarly threshold based on the MSE of the additive term, using a threshold of {$1.0e$--$4$} {on the function variance}.

\color{teal}
\color{black}


\section{Experiments}

We compare across multiple synthetic datasets obeying the additive noise model (ANM) while varying the degree of the additive models.
Following previous works
\citep{buhlmann2014CAM,peters2014continuousANM,rolland2022SCORE_leafDetectionAlgorithm,andrews2023fastBOSS}, we generate the base DAG from an Erdos-Renyi random graph, using an average of 4 edges per node.
We generate 1D, 2D, and 3D additive models to distinguish different hypergraphical structures.
For 1D models, we avoid the linear model to allow for identifiability and use a random Gaussian process to define the additive functions.
For 2D and 3D models, we use multilinear terms, $\beta_{jk} x_j x_k$, with coefficients drawn around $\pm1$,
then normalizing by the total coefficient weight for a parent set.
We assume the additive noise terms are coming from a Gaussian distribution and draw random variances constrained to be close to $1.0$.
We set $d=30$ (number of nodes) and $n=10000$ (number of samples) in our experiments.

We compare against baselines of PC \citep{spirtes1991pc}, GES \citep{chickering2002optimalGES}, BOSS \citep{andrews2023fastBOSS}, RESIT \citep{peters2014continuousANM}, 
CAM \citep{buhlmann2014CAM}, and SCORE \citep{rolland2022SCORE_leafDetectionAlgorithm}.
We additionally compare against a baseline which assumes the empty graph, equivalent to assuming all of the observed variables are completely indepedent.
We compare against the structural Hamming distance (SHD) and the structural interventional distance (SID) obtained by each method on different complexities of synthetic data.
We additionally compare the Hamming distance on the hypergraph bodies, based on the assumption in Figure \ref{fig:hypergraph_teaser}f, calling it the higher-order structural Hamming distance.
Because the maximal error is on the order $2^{30} \approx$ one billion, in some cases, we do not compute exactly and report a lower bound in Table \ref{tab:HOSHD_results_table}.


\subsection{Results}

\begin{table*}
    \centering
    \caption{ {Structural Hamming Distance for ER4 and N=10,000.} }
    \label{tab:SHD_results_table}
\resizebox{\textwidth}{!}{%
    \begin{tabular}{rllllllll}
      \toprule 
                    & \bfseries BOSS      & \bfseries CAM       & \bfseries GES       & \bfseries PC        & \bfseries SCORE     & \bfseries RESIT     & \bfseries zero      & \bfseries HCAM    \\ 
ER4 1D        & 189.67$\pm$20.24    & \textbf{\phantom{0}42.67$\pm$4.50}      & 206.67$\pm$11.90    & 159.67$\pm$21.17    & \textbf{\phantom{0}69.33$\pm$20.24 }      &   494.67$\pm$3.30   & 128.67$\pm$8.38    &  \textbf{{\phantom{0}67.67$\pm$13.82}}   \\
ER4 2D        & 115.33$\pm$\phantom{0}1.70     & 134.67$\pm$2.49     & 116.00$\pm$\phantom{0}2.16     & 134.00$\pm$\phantom{0}4.32     & 126.00$\pm$\phantom{0}2.83     & 146.33$\pm$5.79     &  {115.33$\pm$1.70}     & \textbf{107.00$\pm$\phantom{0}0.82}     \\
ER4 3D        & 109.00$\pm$\phantom{0}4.24     & 120.33$\pm$1.89     &  106.67$\pm$\phantom{0}3.86     & 120.33$\pm$\phantom{0}3.86     & 117.00$\pm$\phantom{0}5.89     & 132.33$\pm$6.34     &  \textbf{106.67$\pm$3.86}     & 106.67$\pm$\phantom{0}3.86     \\
      \bottomrule 
    \end{tabular}
    }
\end{table*}

\begin{table*}
    \centering
    \caption{ {Structural Intervention Distance for ER4 and N=10,000.} }
    \label{tab:SID_results_table}
\resizebox{\textwidth}{!}{%
    \begin{tabular}{rllllllll}
      \toprule 
                    & \bfseries BOSS      & \bfseries CAM       & \bfseries GES       & \bfseries PC        & \bfseries SCORE     & \bfseries RESIT     & \bfseries zero      & \bfseries HCAM    \\ 
ER4 1D        & 608.33$\pm$45.09    &  {\textbf{\phantom{00}0.00$\pm$\phantom{0}0.00}} & 646.67$\pm$54.97    & 748.33$\pm$66.04    &\textbf{105.33$\pm$47.12}   &  646.67$\pm$54.97   & 726.00$\pm$64.19  &  \textbf{525.33$\pm$64.25} \\
ER4 2D        & 679.33$\pm$38.94    & 750.67$\pm$\phantom{0}7.93     & 687.33$\pm$39.67    & 713.67$\pm$60.18    & 745.67$\pm$16.11    & 734.67$\pm$21.64    &  \textbf{679.33$\pm$38.94}    &  \textbf{661.00$\pm$31.03}    \\
ER4 3D        &  \textbf{647.67$\pm$15.08}    & 744.67$\pm$22.23    & 679.00$\pm$39.02    & 699.00$\pm$46.31    & 751.67$\pm$24.14    & 744.00$\pm$29.44    &  \textbf{679.00$\pm$39.02}    & 679.00$\pm$39.02    \\
      \bottomrule 
    \end{tabular}
    }
\end{table*}

\begin{table*}[!t]
    \centering
    \caption{ {Higher-Order Structural Hamming Distance for ER4 and N=10,000.} }
    \label{tab:HOSHD_results_table}
\resizebox{\textwidth}{!}{%
    \begin{tabular}{rllllllll}

      \toprule 
                    & \bfseries BOSS      & \bfseries CAM       & \bfseries GES       & \bfseries PC        & \bfseries SCORE     & \bfseries RESIT     & \bfseries zero      & \bfseries HCAM    \\
ER4 1D        &  >10,000  &    \textbf{\phantom{0}42.67$\pm$\phantom{0}4.50}        &  >10,000   &  >1,000  & >10,000  &  >10,000  & 128.67$\pm$\phantom{0}8.38     & \textbf{\phantom{0}48.33$\pm$\phantom{0}8.18}      \\
ER4 2D        & 157.33$\pm$\phantom{0}1.70     & 168.00$\pm$\phantom{0}5.35     & 158.00$\pm$\phantom{0}1.41     & 183.67$\pm$10.78    & 171.33$\pm$\phantom{0}2.05     & {602.33$\pm$603.66}   & 157.33$\pm$\phantom{0}1.70     & \textbf{119.00$\pm$\phantom{0}5.35}     \\
ER4 3D        & 236.00$\pm$18.38    & 256.33$\pm$14.66    & 248.67$\pm$18.15    & 264.67$\pm$23.23    & 264.00$\pm$14.76    & 263.00$\pm$\phantom{0}13.44    &  \textbf{248.67$\pm$18.15}    & 248.67$\pm$18.15    \\
      \bottomrule 
    \end{tabular}
    }
\end{table*}

%

In Tables \ref{tab:SHD_results_table}, \ref{tab:SID_results_table}, \ref{tab:HOSHD_results_table},
we show the results of each learning algorithm under the SHD, SID, and HO-SHD metrics.
We bold those methods which have statistically significant performance compared with the simple baseline.
Overall, we find that many methods are successful for the simpler 1D data obeying the CAM assumptions,
but struggle on the 2D and 3D datasets.
On the simplest 1D data,
the CAM and SCORE algorithms perform especially well
and
HCAM also achieves good performance on this dataset.
Surprisingly, all other methods have rather great difficulty in identifying the causal structure.

On our specifically higher-order datasets, however,
we find that the story is quite different.
The only algorithm which is able to defeat the baseline on the 2D data is our HCAM method specifically focusing on modeling the 2D interactions.
In the 3D data, none of the algorithms we run are able to find empirical success over the zero baseline.
That is to say, we should have just assumed the variables were independent and not run our algorithm at all.
In several cases, BOSS on 2D, and BOSS, GES, HCAM on 3D, we find that the performance of the baseline is matched by directly predicting the zero baseline (partially or completely).

This lack of capability is despite the fact that we used a simple DGP (multilinear plus Gaussian) on a relatively low number of variables ($d=30$) and provide a relatively standard number of observations ($n=10,000$).
Aligning with our hypothesis that the statistical complexity increases in the presence of higher-order interactions,
we strongly believe this points to some aspects of structure discovery research which have received less attention but remain highly influenced by the presence of interactions.

\section{Conclusion}
We have introduced a framework for considering the impact of higher-order interactions on causal structure.
After introducing the relevant definitions, we further show the identifiability of the introduced structure across multiple cases of interest.
Finally, we demonstrate the potential usefulness of the hypergraphical structure in empirical case using the additive noise model, along with providing a first algorithm for adequately handling the higher-order structure directly from observed data.
This perspective additionally allowed us to identify a potential blindspot of many existing structure discovery approaches:
their lack of focus on statistical power and lesser ability to handle higher-order interactions.

We envision future work may continue to benefit from a perspective using higher-order interactions.
Some directions of future exploration include improving on the algorithms and theoretical results presented herein,
potentially solving increasingly challenging datasets constructed using the higher-order perspective on the generated variables.
Extension to appropriately handling latent variables and latent confounding is a direction of serious interest.
Hypergraphical structures being identifiable from the data distribution alone,
extending existing MEC and identifiability results, point to the potential of hypergraphical structure across even more contexts and settings than the ones explored herein.



\newpage





\bibliography{refs}
\bibliographystyle{plainnat} 

\newpage

\onecolumn

\title{Higher-Order Causal Structure Learning in Additive Noise Models\\(Supplementary Material)}
\maketitle

\appendix

\section{New Definitions}
\label{app_sec:full_definitions}

Due to the introduction of several new definitions because of the need to the classical theory to hypergraphs as in Theorem 1, we dedicate this section for an extended version of Table \ref{tab:notation_table}, providing all definitions in the same place.

\bigskip
\begin{definition}
    (\textbf{Skeleton}) $\quad$ 
    The skeleton of a directed graph $\cG$ is the undirected graph $\cG'$ which has the same vertex set but removes the orientation from each of its edges.  
    If we have the directed edge set $E$, then the edge set of the skeleton is $E'=\{ \{i,j\} : (i,j) \in E\}$.
\end{definition}

\begin{definition}
    (\textbf{Body}) $\quad$ 
    The body of a directed hypergraph $\cH$ is the undirected hypergraph $\cH'$ which has the same vertex set but removes the orientation from each of its hyperedges.
    If we have the directed hyperedge set $H$, 
    then the hyperedge set of the body is $H’ = \{ S+\{ j \} : (S,j) \in H\}$.
\end{definition}

\bigskip
\begin{definition}
    (\textbf{Collider}) $\quad$ 
    A triple of vertices $(i,j,k)$, or more accurately a pair of parent vertices and a single child vertex $((i,k),j)$, are a collider in the directed graph $\cG$ if both $i$ and $k$ are parents of the child node $j$.  
    It is called as such due to the collision of the two arrows coming from nodes $i$ and $k$ at node $j$.
    A collider is further called unshielded if there is no edge between $i$ and $k$.
    This means that $(i,k) \notin E$ and $(k,i) \notin E$.
\end{definition}

\begin{definition}
    (\textbf{Multi-Collider}) $\quad$ 
    A set of parents and a child, $(S,j)$, are called a multi-collider (of order $|S|$) in the directed hypergraph $\cH$ if every $s\in S$ is a parent of the child node $j$.  
    This can be equally written as $(s,j)\in H$ for all $s\in S$.
    Please note that this does not require that $(S,j) \in H$.
    A multi-collider is further called unshielded if there is no hyperedge jointly connecting $S$.
    This means that $(S-t,t) \notin H$ for all $t\in S$.
\end{definition}

\bigskip
\begin{definition}
    (\textbf{Moralized Graph}) $\quad$ 
    The moralized version of a directed graph $\cG$ is an undirected graph $\cG'$ which has the same vertex set as $\cG$.
    In addition to the undirected edges in the skeleton, we also add an undirected edge corresponding to any two vertices $i$ and $k$ which are coparents to a vertex $j$.
    It can be seen that every edge which is added by 'moralization' corresponds to an unshielded collider.
\end{definition}

\begin{definition}
    (\textbf{Immoralized Hypergraph}) $\quad$ 
    The immoralized version of a directed hypergraph $\cH$ is an undirected hypergraph on the same vertex set as $\cH$.
    In addition to the undirected hyperedges in the body, we also add an undirected hyperedge for any set of vertices $S$ which are all coparents to a vertex $j$.
    Again, this does not require that $(S,j) \in H$.
    It can again be seen that every hyperedge which is added by 'immoralization' in this way corresponds to an unshielded multicollider.
    The supposed morality of marrying each pair of parents is contrasted with the supposed immorality of marrying all parents simultaneously.
\end{definition}

\bigskip
\begin{definition}
    (\textbf{Conditional Independence}) $\quad$ 
    In this work where we assume positive density for simplicity, independence between two variables occurs if and only if $\log p(x_i, x_j) = \theta_i(x_i) + \theta_j(x_j)$ for some arbitrary functions $\theta_i$ and $\theta_j$.
    Extension to zero density is fairly straightforward by extending the domain from $(-\infty, \infty)$ to $[-\infty,\infty)$.
    Conditional independence occurs similarly iff $\log p(x_i, x_j | z) = \theta_i(x_i; z) + \theta_j(x_j; z) + \cZ(z)$ for some arbitrary functions $\theta_i$ and $\theta_j$ and a renormalizing function $\cZ(z)$.
\end{definition}
\begin{definition}
    (\textbf{Conditional Multi-Independence}) $\quad$ 
    We say that three variables are 3-independent (or more generally multi-independent) if $\log p(x_i, x_j, x_k) = \theta_{ij}(x_i,x_j) + \theta_{ik}(x_i,x_k) + \theta_{jk}(x_j,x_k)$. for some arbitrary functions $\theta_{ij}$, $\theta_{ik}$, and $\theta_{jk}$.
    Conditional multi-independence occurs similarly iff $\log p(x_i, x_j | z) = 
    \theta_{ij}(x_i,x_j, x_k; z) + \theta_{ik}(x_i,x_k; z) + \theta_{jk}(x_j,x_k; z) + \cZ(z)$ for some arbitrary $\theta_{ij}$, $\theta_{ik}$, $\theta_{jk}$ and a renormalizing  $\cZ(z)$.
    Note that the wording here is inspired more by the case of dependence, that is to say it is more intuitive to say three variables are 3-dependent if they are not only pairwise dependent, but also jointly (multi) dependent as a triple.
\end{definition}





\bigskip
\begin{definition}
    (\textbf{Markov Equivalence}) $\quad$ 
    We say two directed acyclic graphs are Markov equivalent if they induce the same conditional independences.  
    We also say the two directed acyclic graphs are Markov equivalent if they have the same skeletons and unshielded colliders.
    The fact that these two definitions are the same was an important mark of progress for graphical methods originally requiring careful distinction between the two; however,
    now the terms are used more or less interchangeably.
\end{definition}
\begin{definition}
    (\textbf{Hyper-Markov Equivalence}) $\quad$ 
    We say two directed acyclic hypergraphs are hyper Markov equivalent if they induce the same conditional multi-independences.  
    We also say the two directed acyclic hypergraphs are hyper Markov equivalent if they have the same bodies and unshielded multicolliders.
    The fact that these two definitions is the claim of Theorem \ref{thm:hyper_MECs}.
\end{definition}

\section{Proofs of Theorems}

\subsection{Proof of Theorem \ref{thm:hyper_MECs}}

To begin, we remind that while the classical MEC formulation is concerned with mapping the conditional independencies of a distribution with the Markov conditions and structure of a DAG,
we are here concerned with mapping the conditional multi-independencies of a distribution with the Markov conditions and structure of an HDAG.
Accordingly, it is first worth noting that because the conditional multi-independencies of a distribution is a strictly larger set of conditions that the original set of conditions, hence, we get the existing result on the DAG corresponding to the HDAG, let's call it the reduced DAG.
This is done in the obvious way by taking the union of all a node's hyperparents and defining them as the parent set.

Thus,
in addition to the skeletons and (unshielded) colliders of the DAG which are already identifiable from the typical conditions, we must investigate which HDAGs are further distinguishable via these conditions and which HDAGs are distinguishable via the new conditions.

We will start with the easier point about the body of an HDAG.
Once again, this is defined by removing all directed arrows from the directed hypergraph.
In the same way that a pair of nodes $i$ and $j$ are `inseparable' if they are not conditionally indepedent for any conditioning set, 
we can say that a triple of nodes $i$, $j$, and $k$ are inseparable if they are not conditionally multi-independent for any conditioning set.
In much the same way this indicates the existence of an edge in the DAG case, this will indicate the existence of a (three-dimensional) hyperedge in the HDAG case.

It is thus straightforward to see that the existence of an inseparable triple shows the existence of a directed hyperedge (where one of the three vertices is the child).
Further, this generalizes to all degrees in the exact same way.
It is briefly reminded that the hierarchy constraint plays a role of convenience here in the sense that a higher-order edge is detected via a three-dimensional hyperedge even if, say, the generating equations do not make this explicit.
This exactly parallels what happens in the 2D case with an inseparable pair of nodes, where the DAG edges are capturing everything 'between $i$ and $j$ or higher'.
To be explicit, the DAG edge $i\to j$ could have a second parent of $j$ which interacts with $i$ when generating $j$.
Nonetheless, it is clear from these conditions that we may directly identify the body of the HDAG, and that the body of the HDAG is strictly more informative than the skeleton of the HDAG's reduced DAG.


Now let us move on to the discussion of colliders between parents.
To prepare for our generalization of colliders, we first allude to the fact that in Equation \ref{eqn:classical_DAG_Markov_Z_score}, we can see directly that the normalizing $\cZ$ score over the parent set is the cause of a collider.
In particular, unlike the natural $\theta$ terms which cannot be destroyed via marginalization of variables, the $\cZ$ terms are destructible under marginalization of the child.
This naturally corresponds to the more typical conditions noting that there is some smaller set (not including the child) where conditioning provides independence but conditioning on the child additionally breaks the independence.
Of course, not all sets without the child included is able to marginalize out the child, namely, conditioning on any descendant of the child is also problematic.

Nonetheless, we may proceed by extending the definition in the same way.
We say that a set of nodes $i$, $j$, $k$ alongside a fourth node $\ell$ are a tricollider so long as there exists some conditioning set $S$ under which the $i$, $j$, $k$ are not conditionally tri-dependent; however, after additionally conditioning on the node $\ell$ (their joint child), 
the tri-independence breaks and $i$, $j$, $k$ are conditionally tri-dependent when conditioning on $(S + \{\ell\})$.
Equally, it can be seen that $i$, $j$, $k$ are the joint parents of $\ell$ whose $\cZ$ normalization appears only when needing to condition on $\ell$.

In fact, it is now extremely straightforward to state our faithfulness condition directly.
We say that a distribution observes a hypergraph structure faithfully so long as no natural theta term is destructible via marginalization; moreover, the joining of two theta terms via marginalization will lead to a new theta term with the maximal relative size (i.e. no higher-order terms magically cancel and zero out).

Finally,
because of this faithfulness condition,
we can equally start with the log-probability score function which obeys the Hyper Markov property and construct all possible multi-dependence tests in the backwards direction.
Accordingly, no $\theta$ term will cancel via marginalization and each $\cZ$ will only cancel via marginalization of its respective child.
As a reminder,
conditioning on a set is easier via implicitly pulling out the variables value in the conditional distribution,
and marginalizing has been made possible via the faithfulness distribution.
Together, these allow us to describe all of the energy terms of a new conditioned distribution and one can directly read off the conditional multi-dependence via the existence of or lack of the highest-order energy term.


\subsection{Proof of Theorem \ref{thm:thm1}}

\begin{proof}

The arguments here closely follow the original arguments for Theorem 1 of \cite{hoyer2008nonlinearCDwithAdditiveNoiseModels}.

First, recall that we will write
\begin{align}
    \nonumber
    \pi(x_{[d]},y) &= \nu( y - f(x) ) + \xi(x) \\
    \nonumber
    \pi(x_{[d]},y) &= \tilde{\nu}( x_i - g(x_{-i},y) ) + \eta(x_{-i},y) \\
\end{align}
We may first proceed with some basic calculations
\begin{align}
    \nonumber
    \frac{\partial}{\partial x_i} \pi &= \nu' \cdot -\frac{\partial f}{\partial x_i} + 
    \frac{\partial \xi}{\partial x_i} \\
    \nonumber
    &= \tilde{\nu}' \cdot 1 + 0 \\
    \nonumber
    \frac{\partial}{\partial y} \pi &= \nu' \cdot 1 + 0 \\
    \nonumber
    &= \tilde{\nu}' \cdot  -\frac{\partial g}{\partial y}  + 
    \frac{\partial \eta}{\partial y}
\end{align}
And further
\begin{align}
    \nonumber
    \frac{\partial^2}{\partial x_i^2} \pi &= \nu'' \cdot \frac{\partial f}{\partial x_i}\cdot \frac{\partial f}{\partial x_i} - \nu' \cdot \frac{\partial^2 f}{\partial x_i^2} +
    \frac{\partial^2 \xi}{\partial x_i^2} \\
    \nonumber
    &= \tilde{\nu}'' \\
    \nonumber
    \frac{\partial^2}{\partial x_i \partial y} \pi &= \nu'' \cdot -\frac{\partial f}{\partial x_i} \\
    \nonumber
    &= \tilde{\nu}'' \cdot 1  \cdot  -\frac{\partial g}{\partial y}  + \tilde{\nu}' \cdot 0 +
    0 = -\tilde{\nu}'' \cdot \frac{\partial g}{\partial y} 
\end{align}

So it follows from the $\tilde{\nu}$ equation that 
\begin{align}
    \nonumber
    \frac{   \frac{\partial^2 \pi}{\partial x_i^2}   }{  \frac{\partial^2 \pi}{\partial x_i \partial y}   } = \frac{\tilde{\nu}''}{-\tilde{\nu}'' \cdot \frac{\partial g}{\partial y} }  = \frac{-1}{\frac{\partial g}{\partial y} } 
\end{align}
And further
\begin{align}
    \nonumber
     \frac{\partial}{\partial x_i}  \Big[ \frac{   \frac{\partial^2 \pi}{\partial x_i^2}   }{  \frac{\partial^2 \pi}{\partial x_i \partial y}   } \Big]  = 
     \frac{\partial}{\partial x_i} \Big[\frac{-1}{\frac{\partial g}{\partial y} }  \Big] = 0
\end{align}
Plugging this back in to the equation with $\nu$ gives us 
\begin{align}
    \nonumber
     \frac{\partial}{\partial x_i}  \Big[ \frac{   \nu''\cdot f' \cdot f' - \nu' f'' + \xi''  }{  -\nu'' f'   } \Big]  \equiv 0
\end{align}
Application of repeated derivative rules and simplification gives the required
    \begin{align}
    \nonumber
        \xi''' = \xi'' \cdot \Big(-\frac{\nu''' f'}{\nu''} + \frac{f''}{f'} \Big) + \\
        \Big(-2\nu'' f'' f' + \nu' f''' + \frac{\nu' \nu''' f' f''}{\nu''} - \frac{\nu' f'' f''}{f'}    \Big) \nonumber
    \end{align}
\end{proof}


\subsection{Proof of Theorem \ref{thm:thm2}}

\begin{proof}
    Supposing that we have two different forward models given by 
        \begin{align}
        p(x_1,\dots,x_d,y) = p_n(y - \sum_{S\in\cI} f_S(x_S)) \cdot p_x(x_1,\dots,x_d) 
         = p_n(y - f(x) ) \cdot p_x(x_1,\dots,x_d) 
         \nonumber
    \end{align}
    \begin{align}
        p(x_1,\dots,x_d,y) = p_{\Tilde{n}}(y - \sum_{T\in\cJ} g_T(x_T)) \cdot p_x(x_1,\dots,x_d) 
        = p_{\Tilde{n}}(y - g(x) ) \cdot p_x(x_1,\dots,x_d) 
        \nonumber
    \end{align}
    We may take $\pi$ as before and see that
    \begin{align}
        \nonumber
        \frac{\partial}{\partial y} \pi &=& \nu' &=& \tilde{\nu}' \\ 
        \nonumber
        \frac{\partial}{\partial x_i} \pi &=& -\nu' \cdot \Big[  \sum_{S\in\cI} 
        \frac{\partial f_S}{\partial x_i} (x_S) \ind(i\in S) \Big]  &=& -\tilde{\nu}' \cdot \Big[  \sum_{T\in\cJ} 
        \frac{\partial g_T}{\partial x_i} (x_T) \ind(i\in T) \Big]
    \end{align}

    It follows as before that we may write 
    \begin{align}
        \nonumber
        -\Big[ \frac{  \frac{\partial}{\partial x_i} \pi }{  \frac{\partial}{\partial y} \pi }\Big] &=& \Big[  \sum_{S\in\cI} 
        \frac{\partial f_S}{\partial x_i} (x_S) \ind(i\in S) \Big]  &=& \Big[  \sum_{T\in\cJ}         \frac{\partial g_T}{\partial x_i} (x_T) \ind(i\in T) \Big]
    \end{align}

    Further, we have that
    \begin{align}
        \nonumber
        0 \equiv \frac{\partial}{\partial x_{R-i}}
        \Big[ 0 \Big] = 
        \frac{\partial}{\partial x_{R-i}}
        \Big[ \frac{  \frac{\partial}{\partial x_i} \pi }{  \frac{\partial}{\partial y} \pi }   - \frac{  \frac{\partial}{\partial x_i} \pi }{  \frac{\partial}{\partial y} \pi }   \Big]
        =
        \frac{\partial}{\partial x_{R-i}}
        \Big[ - \frac{\partial f}{\partial x_i}  + \frac{\partial g}{\partial x_i}    \Big]
        =
        -\frac{\partial f}{\partial x_{R}}  + \frac{\partial g}{\partial x_{R}}
        \\
        = - \Big[  \sum_{S\in\cI} 
        \frac{\partial f_S}{\partial x_R} (x_S) \cdot \ind(R\subseteq S) \Big]  + 
        \Big[  \sum_{T\in\cJ}         \frac{\partial g_T}{\partial x_R} (x_T) \cdot \ind(R\subseteq T) \Big]
    \end{align}

    Recall that we assumed that $\cI$ and $\cJ$ are downwards closed or `hierarchical' which means all $S\in\cI$ have all its subsets $S'\subseteq S$ also inside of $\cI$.
    If we then take $R\notin\cI$, it implies that $R\not\subseteq S$ for all $S\in\cI$, otherwise such an $S$ would not obey the downwards closed property.
    This means that $\ind(R\subseteq S) = 0$ and
    \begin{align}
        \nonumber
        0 \equiv 
         - 0  + 
        \Big[  \sum_{T\in\cJ}         \frac{\partial g_T}{\partial x_R} (x_T) \cdot \ind(R\subseteq T) \Big]
    \end{align}
    This means that for all $T \in \cJ \cap \text{up}(R)$ where $\text{up}(R) := \{ T : T\supseteq R\}$ it must be the case that the $R$-th partial derivative is zero.
    We will focus on the case of $T=R$, but the same holds for all $T$ as above.

    We may take a modification of the function $g_T$ such that it is instead represented by additive functions of a lower degree.
    This is equivalent to saying that $g_T\equiv 0$ and hence $T\in\cJ$ was actually a contradiction.
    Let us see that $\frac{\partial g_R}{\partial x_R}\equiv 0$, so then writing $R=\{r_1,\dots,r_{|R|}\}$, we have that $\int_{x_{r_1}} \frac{\partial g_R}{\partial x_R} = C_1(x_{r_2},\dots,x_{r_{|R|}})$ for some function $C_1$ which is constant with respect to $x_{r_1}$. Further $\int_{x_{r_2}} \frac{\partial g_{R}}{\partial x_{R-r_1}} = \int_{x_{r_2}} C_1(x_{r_2},\dots,x_{r_{|R|}}) = C_1(x_{r_2},\dots,x_{r_{|R|}}) + C_2(x_{r_1},x_{r_3},\dots,x_{r_{|R|}})$ and continuing on, we may ultimately see that $g_R = \sum_{R'\subsetneq R} g_{R'}$ which means by our assumption that $\cJ$ is a minimal representation with no zero additive models, that actually $T\notin\cJ$.

    The same arguments may be taken in reverse to show that for all $R\not\in\cJ$, it must be the case $R\not\in\cI$.
    There is a mild difference in the way the contradiction is applied argument when reversing the arguments because we take the perspective that $\cI$ is the ground truth generating process and $\cJ$ is a potential alternate model (i.e. the contradiction is on $R$ not on $S$.)
    Nonetheless, this results in the fact $\cI=\cJ$ for any two forward models which are representing the same distribution, and that the functional representations are moreover the same.
    The latter part can be seen directly from $ \frac{\partial f}{\partial x_i}  = \frac{\partial g}{\partial x_i}   $ for all $i\in[d]$ and similar arguments for removing trivial terms from the additive model.
    The final modification which may exist is up to a constant, which is resolved by the differences in the mean of the variable represented by $\nu$ and $\tilde{\nu}$.
    This is solved by the ANM assumption that the additive noises are mean-centered.
\end{proof}

Altogether, this is taken to mean that whenever the DAG is locally identifiable (and thus there are only valid forward models and no potential backwards model), then the hypergraph structure of the forward model is additionally identifiable.
This additionally implies that if the entirety of the DAG is identifiable, then the entirity of the hyper DAG is also identifiable.

\subsection{Proof of Theorem \ref{thm:thm3}}

\begin{proof}
Under the further assumption that $\eps_i \sim \cN(0,\sigma_i^2)$, we have that $\nu(\eps) = \log p(\eps) = \log \big( \frac{1}{\sqrt{2\pi\sigma^2}} \cdot \exp(-\frac{\eps^2}{2\sigma^2}) \big) = -\log(2\pi\sigma^2) - \frac{1}{2\sigma^2} \eps^2 $.
Further, we have that $\nu'(\eps) = - \frac{1}{\sigma^2} \eps$, $\nu''(\eps) = - \frac{1}{\sigma^2}$, and $\nu^{(k)}(\eps) = 0$ for larger k. 
Accordingly, we may write the entire distribution as 
\begin{align}
    \nonumber
    \xi(x_1,\dots,x_d) &= \log p(x_1,\dots,x_d) = \sum_i \log p(x_i | x_{\text{Pa}(i)} ) \\
    \nonumber 
    &=
    \sum_i \nu_i(\eps_i) = \sum_i \nu_i\Big( x_i - \sum_{(S,i)\in\cH} f_{S\rightarrow i}(x_S) \Big)
\end{align}
Let us write $f_{\rightarrow i}$ to denote $\sum_{(S,i)\in\cH} f_{S\rightarrow i}$ and $F_i(x) = (x_i - f_{\rightarrow i}(x) )$.
It is straightforward to verify through repeated applications of the chain rule and product rule that
\begin{align}
    \nonumber
    \frac{\partial^n}{\partial x_1,\dots,\partial x_n} \nu_i\Big( F_i(x) \Big) =    
    \nu_i \cdot 
    \frac{\partial^n}{\partial x_1,\dots,\partial x_n} \Big(F_i(x) \Big) + \nu_i' \cdot  \sum_{\emptyset\subsetneq A\subsetneq[n]} \frac{\partial^{|A|}}{\partial x_A} \Big(F_i(x) \Big) \cdot \frac{\partial^{n-|A|}}{\partial x_{[n]-A}} \Big(F_i(x) \Big)
\end{align}

It can further be seen that 
\begin{align}
    \frac{\partial^{|A|}}{\partial x_A} \Big(F_i(x) \Big)  = 
    \frac{\partial^{|A|}}{\partial x_A} \Big( x_i - f_{\rightarrow i}(x) \Big) 
\end{align}
is zero whenever $A$ is not all $i$'s and $A$ is not a subset of one of the $S$ where $(S,i)\in\cH$.
This means exactly that 
\[
\sum_{\emptyset\subsetneq A\subsetneq[n]} \nu_i' \cdot \frac{\partial^{|A|}}{\partial x_A} \Big(F_i(x) \Big) \cdot \frac{\partial^{n-|A|}}{\partial x_{[n]-A}} \Big(F_i(x) \Big)
\]
is barely nonzero whenever we take $[n]$ equal to $(S+i)$ for some $(S,i)\in\cH$.
Note that this is equivalent to taking some $(S+i)\in\cH'$ where we recall $\cH'$ is the undirected version of the directed graph $\cH$.
If we instead take some $R\notin \cH'$,
then it will be the case that this derivative is zero, because all $A\subseteq R$ will have either $\frac{\partial^{|A|}}{\partial x_A} \equiv 0$ or $\frac{\partial^{n-|A|}}{\partial x_{[n]-A}} \equiv 0$.
Moreover, it is the case that the first term is clearly zero $\frac{\partial^{|R|}}{\partial x_R} \Big(F_i(x) \Big) \equiv 0$.

Since this is true for all $i$ so long as we are taking $R\notin\cH'$,
we have that 
\begin{align}
    \nonumber
    \frac{\partial^{|R|}}{\partial x_R} \xi(x_1,\dots,x_d) = \frac{\partial^{|R|}}{\partial x_R} \sum_i \nu_i\Big( x_i - \sum_{(S,i)\in\cH} f_{S\rightarrow i}(x_S) \Big) \equiv 0    
\end{align}
Following the same approach as in the proof of Theorem \ref{thm:thm2},
this means that we are able to write $\xi(x_1,\dots,x_d) = \sum_{S\in\cH'} \xi_S(x_S)$ for some functions $\xi_S$.

\end{proof}

Note that the decomposition of the likelihood function's structure does not contradict existing results saying that the directionality of the graphical model is not always identifiable from data.
In particular, in the linear-Gaussian case, it may not be possible to distinguish which direction is the causal direction.
Nonetheless, the graphical structure which is recovered in this purely Gaussian case corresponds to what is available in the precision matrix \citep{loh2014linearCausalNetworksViaPrecisionMatrix}, still identifying the undirected graphical structure underlying the distribution.
Under the CAM-like assumptions of linear SEMs, the hypergraph structure is reduced to its simplest representation, which is isomorphic to a graphical representation.

\newpage

\newpage

\section{Experiment Details}

The data is generated as follows.  We first sample the ER graph and then sample a random ordering (it does not matter which order we do this in).  In the 1D setting our HDAG is already complete.  In the 2D and 3D setting, we choose hyperedges by taking a random ordering of parents and selecting pairs or triples in a cyclic fashion.  This is the simplest HDAG which obeys the same DAG structure. Now we sample the parameters.  For the beta coefficients, we sample from [0.5, 2.0] using an log uniform distribution.  We sample gaussian variances from the same distribution. 
We then rescale the coefficients by the square root of the expected Gaussian moments, $\bbV\text{ar}[\eps_1 \eps_2] = 3.0$ and $\bbV\text{ar}[\eps_1 \eps_2 \eps_3] = 15.0$, respectively along with the total number of parents.  For the 1D case we follow the literature standard of sampling from a Gaussian process prior (randomly initialized \texttt{sklearn.gaussian\_process.GaussianProcessRegressor}).


\section{Algorithm Details}
\label{app_sec:algorithm_details}








\subsection{Causal Additive Model}

CAM (Causal Additive Model) uses a three step procedure to discover a set of additive structural equations according to generalized additive model assumption.
First, a preliminary search is made over the directed edges using sparse regression to cut down on the search space.
Second, a greedy algorithm gradually adds the best edges to the DAG so long as it does not create any cycles.
Third, the final DAG structure's additive models are trained once again with sparse regression to encourage the removal of extraneous edges.

\paragraph{Step 1}
First, a preliminary search is made over all possible edges via sparse regression.
For each variable $j\in[d]$, one fits an additive model based on all of the other possible directed edges $(k,j)$ using sparse regression.
This allows for a smaller subset of the quadratic number of edges to be considered, especially in the high-dimensional setting when $d$ is large.

The mean-squared error objective is minimized based on the assumption that the noise terms are Gaussian.
\begin{align}
    \log p(\eps_j) = -\log(2\pi\sigma_j^2) - \frac{1}{2\sigma_j^2} \cdot \eps_j^2
\end{align}
\begin{align}
    \hat{\sigma}_j^2 := \|X_j - \hat{X}_j \|^2
    = \|X_j - \sum_{k} \hat{f}_{k\rightarrow j}(X_k) \|^2
\end{align}

\paragraph{Step 2}
Second,
the bulk of the algorithm centers around a greedy approach for gradually adding directed edges which do not disagree with the partial structure which has been built up so far.
Every edge from the local neighborhood determined in step 1 is considered to be added, so long as it would not create a cycle in the DAG.
Each edge is ranked by its ability to improve the log-likelihood of the overall model, by training an additive model with the selected edges.

\begin{align}
    \hat{\sigma}_j^2(\cN_j) := \|X_j - \sum_{k\in\cN_j} \hat{f}_{k\rightarrow j}(X_k) \|^2
\end{align}
\begin{align}
    (k*,j*) = \argmin_{(k,j) \quad\text{acyclic}} \big\{ \hat{\sigma}_j^2(\cN_j\cup\{k\})  - \hat{\sigma}_j^2(\cN_j) \}
\end{align}

Importantly, for $j\neq j*$, it is not necessary to retrain the additive models to recompute the values of $\hat{\sigma}_j^2(\cN_j)$, because they are not affected by the inclusion of edges in other parts of the graph (except that it may block an edge from being added due to the acyclicity constraint).

\paragraph{Step 3}
Lastly,
the collection of directed edges which were selected in step 2 are used to train a final model end-to-end, with additional regularization designed to shrink unnecessary edges to become sparse.
Additive terms $\hat{f}_{k\rightarrow j}$ which are deemed insignificant are removed from the model completely and the final set of edges define the final DAG.

\subsection{Sparse Interaction Additive Network}

SIAN (Sparse Interaction Additive Network) is an approach designed to train higher-order additive models using neural network techniques.
This approach also consists of three main phases.
In our case, we will follow CAM's implicit Gaussian assumption by minimizing the mean-squared error objective which corresponds with the likelihood of independent Gaussian variables.

In the first phase, a typical neural network $f_\theta$ is trained to predict an output variable in terms of the input variables.

\begin{align}
    \hat{\sigma}^2(\theta) := \|Y - \hat{Y}(\theta) \|^2
    = \|Y - \hat{f}_{\theta}(X_{[d]}) \|^2
\end{align}

In the second phase, interpretability techniques are combined with a
special feature interaction selection (FIS) algorithm which ensures a sufficient coverage of the complex space of interactions while avoiding the exponential blow up in complexity from exploring all higher-order interactions.
The final result of the first two phases is a collection $\cI\subseteq\cP([d])$ which is some collection of all of the feature interactions $S$ which are important to predicting the output variable.
Finally, the set of collected higher-order interactions are then used to train a neural-network-based additive model which obeys the interaction structure determined in the selection algorithm.

\begin{align}
    \hat{\sigma}_{\cI}^2(\theta) :=
    \|Y - \hat{Y}_{\cI}(\theta) \|^2 =
     \|Y - \sum_S \hat{f}_{S,\theta_S}(X_{S}) \|^2
\end{align}

This final additive neural network has pleasant properties like being more interpretable as well as more robust than the original neural network.
In our context, we will use these neural additive models as the major component of modeling the hypergraphical additive structure we assumed previously.



\subsection{Higher-order Causal Additive Model}


{In our algorithm},
we broadly follow the same steps as the original CAM algorithm,
replacing all components which are limited to one-dimensional additive models with their higher-order counterparts.

In the first step of our algorithm, we must reduce the number of candidate edges which will be considered in the downstream steps.
Although CAM mentions this is only necessary in the high-dimensional setting for their additive assumption, ours is absolutely necessary except in extremely small cases (perhaps $d\leq 5$).
This is because instead of searching over all candidate directed edges, $\{(k,j) : k\neq j, j\in[d]\}$,
we must perform a search over the much larger space of all candidate directed hyperedges, $\{ (S,j) : S\subseteq ([d]\setminus j), j\in[d] \}$.

For this purpose, we employ the first two phases of SIAN to each of the variables.
That is, for each $j\in[d]$, we train a neural network to predict $X_j$ from $X_{-j}$ and then run a feature interaction selection algorithm to find a neighborhood of important interactions which are useful for predicting $X_j$.
\begin{align}
    \hat{\sigma}_j^2(\theta) := \|X_j - \hat{X}_j(\theta) \|^2
    = \|X_j - \hat{f}_{j,\theta}(X_{[d]-k}) \|^2
\end{align}
\begin{align}
    \cI_j = \text{FeatureInteractionSelection}(\hat{f}_{j,\theta})
\end{align}

These selected interactions are then taken as the candidate set of directed hyperedges to be used in the later parts of the algorithm.
\begin{align}
    \tilde{\cH} := \{ (S,j) : S\in\cI_j, j\in[d] \}
\end{align}

Note that these hyperedges are also given an importance score from the original FIS algorithm and may be sorted by their a priori importance.

In the second step of our algorithm,
we follow the greedy approach of including hyperedges based on the improvement to the log-likelihood.
This requires training many different additive models obeying the interaction constraints imposed by the current hyper DAG.
We use the additive models from the third phase of SIAN to minimize an MSE objective as before.

In particular, we train multiple SIAN additive models to compare the different improvements in scores coming from adding all possible hyperedges $(S,j)\in\tilde{\cH}$.
This improvement in score is again interpreted as the improvement in reducing the noise of the added Gaussian via reduction in MSE.

\begin{align}
    \hat{\sigma}_j^2(\cN'_j) := \|X_j - \sum_{S\in\cN'_j} \hat{f}_{S\rightarrow j}(X_S) \|^2
\end{align}
\begin{align}
    (S*,j*) = \argmin_{(S,j) \quad\text{acyclic}} \big\{ \hat{\sigma}_j^2(\cN'_j\cup S)  - \hat{\sigma}_j^2(\cN'_j) \}
\end{align}

Because this requires training a large amount of additive models,
we make multiple concessions to allow for a more rapid selection process during step 2 which can otherwise take a significant chunk of the overall algorithm time.
Because of our higher-order neighborhood selection from step 1, it is at least feasible to search over higher-order interactions without facing an exponential blowup in the number of additive models which must be trained.

However, in practice we further reduce the number of additive models we train to a maximum of 10 interactions per each variable $X_j$.
As tuples are selected from the candidate superset $\tilde{\cH}$ to be actually included into the model, additional candidate hyperedges are replenished to be explored in future iterations of the step 2 loop.

Furthermore, instead of training these SIAN additive models until there is no further reduction in MSE, we only train each for five epochs in total.
We find that this gives a strong enough measurement of the performances of the differnt additive models without cutting significantly into the overall time taken.
Moreover, because the heuristic coming from the first two phases of SIAN used in step 1 of our algorithm is generally quite good, an approximate measure of the reduction in score from each hyperedge in step 2 seems to generally be sufficient. 

In the third step of our algorithm,
we again follow the CAM setup and train an end-to-end SIAN model which obeys the structure which was greedily added in step two of the algorithm.
L1 regularization terms are used on each of the shape functions in the additive model to encourage shrinkage in the unnecessary terms of the additive model.
Similar to CAM, unimportant additive terms are thresholded away and removed from the final hyper DAG.
%
%
\begin{align}
\cL(\theta) := \quad\quad\quad\quad \quad\quad\quad\quad \\
    \sum_j \|X_j - \sum_{(S,j)\in\cH} \hat{f}_{S\rightarrow j} \|^2 + 
    \nonumber
    \lambda_1 \sum_{(S,j)\in\cH} |\hat{f}_{S\rightarrow j}|
\end{align}

For comparison against other DAG-based methods,
the hyper DAG is further projected onto its equivalent DAG formulation.

\end{document}

%% file: jam_header.tex
\DeclareSymbolFont{bbold}{U}{bbold}{m}{n}
\DeclareSymbolFontAlphabet{\mathbbold}{bbold}
\newcommand{\ind}{\mathbbold{1}}
\newcommand{\oct}{\hspace{0.5em}}

\let\epsilon\varepsilon
\let\eps\epsilon

\let\phi\varphi
\usepackage{comment}

\usepackage{relsize}

\newcommand{\indep}{\perp\!\!\!\!\perp}

\usepackage{amsmath, amsthm, amssymb}
\theoremstyle{definition}

\usepackage{ifpdf}
\ifpdf
\usepackage[pdftex]{graphicx}
\else
\usepackage[dvips]{graphicx}
\fi
\newcommand{\makeheading}[1]%
        {\hspace*{-\marginparsep minus \marginparwidth}%
         \begin{minipage}[t]{\textwidth}%
                {\large \bfseries #1}\\[-0.15\baselineskip]%
                 \rule{\columnwidth}{1pt}%
         \end{minipage}}
\def\semicolon{;}
\def\applytolist#1{
    \expandafter\def\csname multi#1\endcsname##1{
        \def\multiack{##1}\ifx\multiack\semicolon
            \def\next{\relax}
        \else
            \csname #1\endcsname{##1}
            \def\next{\csname multi#1\endcsname}
        \fi
        \next}
    \csname multi#1\endcsname}

\def\calc#1{\expandafter\def\csname c#1\endcsname{{\mathcal #1}}}
\applytolist{calc}QWERTYUIOPLKJHGFDSAZXCVBNM;
\def\bbc#1{\expandafter\def\csname bb#1\endcsname{{\mathbb #1}}}
\applytolist{bbc}QWERTYUIOPLKJHGFDSAZXCVBNM1;

\theoremstyle{definition}
\newtheorem{theorem}{Theorem}

\newtheorem{remark}{Remark}
\newtheorem{corollary}{Corollary}

\newtheorem{definition}{Definition} 

\usepackage{amsthm}
\usepackage{amsmath}
\usepackage{amssymb}
\usepackage{graphicx}
\usepackage{mathtools}
\usepackage{enumerate}
\usepackage{enumitem}
\usepackage{footnote}
\usepackage{float}
\usepackage{xspace}
\usepackage{multirow}
\usepackage{nicefrac}
\usepackage{wrapfig}
\usepackage{framed}
\usepackage{url}

\DeclareMathOperator*{\argmin}{argmin}

\usepackage{color}
\definecolor{deepblue}{rgb}{0,0,0.5}
\definecolor{deepred}{rgb}{0.6,0,0}
\definecolor{deepgreen}{rgb}{0,0.5,0}
\definecolor{deeporange}{rgb}{0.8,0.4,0}

\newboolean{showcomments}
\setboolean{showcomments}{true} 
\ifthenelse{\boolean{showcomments}}{
  \newcommand{\nbc}[3]{
    {\colorbox{#3}{\bfseries\sffamily\scriptsize\textcolor{white}{#1}}}%
    {\textcolor{#3}{\textsf\small$\blacktriangleright$\textit{#2}$\blacktriangleleft$}}}
  \newcommand{\todo}[1]{\nbc{TODO}{#1}{blue}\xspace}
}{
  \newcommand{\nbc}[3]{}
  \renewcommand{\todo}[1]{}
}

\usepackage{multicol}

\renewcommand\vec[1]{\ifstrequal{#1}{0}{\ensuremath{\mathbf{0}}}{\ensuremath{\boldsymbol{#1}}}}

%% file: neurips_main_arxiv.bbl
\begin{thebibliography}{44}
\providecommand{\natexlab}[1]{#1}
\providecommand{\url}[1]{\texttt{#1}}
\expandafter\ifx\csname urlstyle\endcsname\relax
  \providecommand{\doi}[1]{doi: #1}\else
  \providecommand{\doi}{doi: \begingroup \urlstyle{rm}\Url}\fi

\bibitem[Amari et~al.(2003)Amari, Nakahara, Wu, and Sakai]{amari2003higherOrderNeurons}
Shun-Ichi Amari, Hiroyuki Nakahara, Si~Wu, and Yutaka Sakai.
\newblock Synchronous firing and higher-order interactions in neuron pool.
\newblock \emph{Neural Computation}, 2003.

\bibitem[Andersson et~al.(1997)Andersson, Madigan, and Perlman]{andersson1997characterizationOfMECs}
Steen~A. Andersson, David Madigan, and Michael~D. Perlman.
\newblock A characterization of markov equivalence classes for acyclic digraphs.
\newblock \emph{The Annals of Statistics}, 25\penalty0 (2):\penalty0 505--541, 1997.
\newblock ISSN 00905364, 21688966.
\newblock URL \url{http://www.jstor.org/stable/2242556}.

\bibitem[Andrews et~al.(2023)Andrews, Ramsey, Romero, Camchong, and Kummerfeld]{andrews2023fastBOSS}
Bryan Andrews, Joseph Ramsey, Ruben~Sanchez Romero, Jazmin Camchong, and Erich Kummerfeld.
\newblock Fast scalable and accurate discovery of {DAG}s using the best order score search and grow shrink trees.
\newblock In \emph{Thirty-seventh Conference on Neural Information Processing Systems}, 2023.
\newblock URL \url{https://openreview.net/forum?id=80g3Yqlo1a}.

\bibitem[Bartlett and Cussens(2017)]{bartlett2017integer}
Mark Bartlett and James Cussens.
\newblock Integer linear programming for the {Bayesian} network structure learning problem.
\newblock \emph{Artificial Intelligence}, 244:\penalty0 258--271, 2017.

\bibitem[Battiston et~al.(2020)Battiston, Cencetti, Iacopini, Latora, Lucas, Patania, Young, and Petri]{battiston2020networksBeyondPairwiseInteractions}
Federico Battiston, Giulia Cencetti, Iacopo Iacopini, Vito Latora, Maxime Lucas, Alice Patania, Jean-Gabriel Young, and Giovanni Petri.
\newblock Networks beyond pairwise interactions: Structure and dynamics.
\newblock \emph{Physics Reports}, 874:\penalty0 1--92, 2020.
\newblock ISSN 0370-1573.
\newblock \doi{https://doi.org/10.1016/j.physrep.2020.05.004}.
\newblock URL \url{https://www.sciencedirect.com/science/article/pii/S0370157320302489}.
\newblock Networks beyond pairwise interactions: Structure and dynamics.

\bibitem[B{\"u}hlmann et~al.(2014)B{\"u}hlmann, Peters, and Ernest]{buhlmann2014CAM}
Peter B{\"u}hlmann, Jonas Peters, and Jan Ernest.
\newblock {CAM: Causal additive models, high-dimensional order search and penalized regression}.
\newblock \emph{The Annals of Statistics}, 42\penalty0 (6):\penalty0 2526 -- 2556, 2014.
\newblock \doi{10.1214/14-AOS1260}.
\newblock URL \url{https://doi.org/10.1214/14-AOS1260}.

\bibitem[Chickering(2002)]{chickering2002optimalGES}
David~M. Chickering.
\newblock Optimal structure identification with greedy search.
\newblock \emph{Journal of Machine Learning Research}, 3\penalty0 (Nov):\penalty0 507--554, 2002.

\bibitem[Chickering(1995)]{chickering1995equivalentBayesianNets}
David~Maxwell Chickering.
\newblock A transformational characterization of equivalent bayesian network structures.
\newblock In \emph{Proceedings of the Eleventh Conference on Uncertainty in Artificial Intelligence}, UAI'95, page 87–98, San Francisco, CA, USA, 1995. Morgan Kaufmann Publishers Inc.
\newblock ISBN 1558603859.

\bibitem[de~Arruda et~al.(2020)de~Arruda, Petri, and Moreno]{deArruda2020introductionHOmotivation_socialContagionHypergraphs}
Guilherme~Ferraz de~Arruda, Giovanni Petri, and Yamir Moreno.
\newblock Social contagion models on hypergraphs.
\newblock \emph{Phys. Rev. Res.}, 2:\penalty0 023032, Apr 2020.
\newblock \doi{10.1103/PhysRevResearch.2.023032}.
\newblock URL \url{https://link.aps.org/doi/10.1103/PhysRevResearch.2.023032}.

\bibitem[Enouen and Liu(2022)]{enouen2022sian}
James Enouen and Yan Liu.
\newblock Sparse interaction additive networks via feature interaction detection and sparse selection.
\newblock In \emph{Advances in Neural Information Processing Systems}, 2022.

\bibitem[Enouen and Sugiyama(2024)]{enouen2024completeDecompositionfKLerror}
James Enouen and Mahito Sugiyama.
\newblock A complete decomposition of kl error using refined information and mode interaction selection, 2024.
\newblock URL \url{https://arxiv.org/abs/2410.11964}.

\bibitem[Evans(2016)]{evans2016hypergraphsForMarginsOfBayesNets}
Robin~J. Evans.
\newblock Graphs for margins of bayesian networks.
\newblock \emph{Scandinavian Journal of Statistics}, 43\penalty0 (3):\penalty0 625--648, 2016.
\newblock \doi{https://doi.org/10.1111/sjos.12194}.
\newblock URL \url{https://onlinelibrary.wiley.com/doi/abs/10.1111/sjos.12194}.

\bibitem[Freeman and White(1993)]{freeman1993introductionHOmotivation_socialCliques}
Linton~C. Freeman and Douglas~R. White.
\newblock Using galois lattices to represent network data.
\newblock \emph{Sociological Methodology}, 23:\penalty0 127--146, 1993.
\newblock ISSN 00811750, 14679531.
\newblock URL \url{http://www.jstor.org/stable/271008}.

\bibitem[Gaudelet et~al.(2018)Gaudelet, Malod-Dognin, and Pržulj]{gaudelet2018introductionHOmotivation_PPIsAndHypergraphlets}
Thomas Gaudelet, Noël Malod-Dognin, and Nataša Pržulj.
\newblock Higher-order molecular organization as a source of biological function.
\newblock \emph{Bioinformatics}, 34\penalty0 (17):\penalty0 i944--i953, 09 2018.
\newblock ISSN 1367-4803.
\newblock \doi{10.1093/bioinformatics/bty570}.
\newblock URL \url{https://doi.org/10.1093/bioinformatics/bty570}.

\bibitem[Glymour et~al.(2019)Glymour, Zhang, and Spirtes]{glymour2019review}
Clark Glymour, Kun Zhang, and Peter Spirtes.
\newblock Review of causal discovery methods based on graphical models.
\newblock \emph{Frontiers in Genetics}, 10, 2019.

\bibitem[Hoyer et~al.(2008)Hoyer, Janzing, Mooij, Peters, and Sch\"{o}lkopf]{hoyer2008nonlinearCDwithAdditiveNoiseModels}
Patrik Hoyer, Dominik Janzing, Joris~M Mooij, Jonas Peters, and Bernhard Sch\"{o}lkopf.
\newblock Nonlinear causal discovery with additive noise models.
\newblock In D.~Koller, D.~Schuurmans, Y.~Bengio, and L.~Bottou, editors, \emph{Advances in Neural Information Processing Systems}, volume~21. Curran Associates, Inc., 2008.
\newblock URL \url{https://proceedings.neurips.cc/paper_files/paper/2008/file/f7664060cc52bc6f3d620bcedc94a4b6-Paper.pdf}.

\bibitem[Immer et~al.(2023)Immer, Schultheiss, Vogt, Sch{\"o}lkopf, B{\"u}hlmann, and Marx]{immer2023identifiability}
Alexander Immer, Christoph Schultheiss, Julia~E Vogt, Bernhard Sch{\"o}lkopf, Peter B{\"u}hlmann, and Alexander Marx.
\newblock On the identifiability and estimation of causal location-scale noise models.
\newblock In \emph{International Conference on Machine Learning}, 2023.

\bibitem[Javidian et~al.(2020)Javidian, Wang, Lu, and Valtorta]{javidian2020hypergraphPGM}
Mohammad~Ali Javidian, Zhiyu Wang, Linyuan Lu, and Marco Valtorta.
\newblock On a hypergraph probabilistic graphical model.
\newblock \emph{Annals of Mathematics and Artificial Intelligence}, 2020.

\bibitem[Koller and Friedman(2009)]{koller2009probabilisticGraphicalModels}
D.~Koller and N.~Friedman.
\newblock \emph{Probabilistic Graphical Models: Principles and Techniques}.
\newblock Adaptive computation and machine learning. MIT Press, 2009.
\newblock ISBN 9780262013192.

\bibitem[Lachapelle et~al.(2020)Lachapelle, Brouillard, Deleu, and Lacoste-Julien]{lachapelle2020granDAG}
Sébastien Lachapelle, Philippe Brouillard, Tristan Deleu, and Simon Lacoste-Julien.
\newblock Gradient-based neural dag learning.
\newblock In \emph{International Conference on Learning Representations}, 2020.
\newblock URL \url{https://openreview.net/forum?id=rklbKA4YDS}.

\bibitem[Loh and B{{\"u}}hlmann(2014)]{loh2014linearCausalNetworksViaPrecisionMatrix}
Po-Ling Loh and Peter B{{\"u}}hlmann.
\newblock High-dimensional learning of linear causal networks via inverse covariance estimation.
\newblock \emph{Journal of Machine Learning Research}, 15\penalty0 (88):\penalty0 3065--3105, 2014.
\newblock URL \url{http://jmlr.org/papers/v15/loh14a.html}.

\bibitem[Majhi et~al.(2022)Majhi, Perc, and Ghosh]{majhi2022dynamicsOnHigherOrderNetworksAReview}
Soumen Majhi, Matjaž Perc, and Dibakar Ghosh.
\newblock Dynamics on higher-order networks: a review.
\newblock \emph{Journal of The Royal Society Interface}, 19\penalty0 (188), Mar 2022.
\newblock ISSN 1742-5662.
\newblock \doi{10.1098/rsif.2022.0043}.
\newblock URL \url{http://dx.doi.org/10.1098/rsif.2022.0043}.

\bibitem[Ng et~al.(2022)Ng, Zhu, Fang, Li, Chen, and Wang]{ng2022masked}
Ignavier Ng, Shengyu Zhu, Zhuangyan Fang, Haoyang Li, Zhitang Chen, and Jun Wang.
\newblock Masked gradient-based causal structure learning.
\newblock In \emph{SIAM International Conference on Data Mining (SDM)}, 2022.

\bibitem[Pearl(2009)]{pearl2009causalityBook}
Judea Pearl.
\newblock \emph{Causality}.
\newblock Cambridge University Press, 2 edition, 2009.

\bibitem[Peters et~al.(2014)Peters, Mooij, Janzing, and Sch{{\"o}}lkopf]{peters2014continuousANM}
Jonas Peters, Joris~M. Mooij, Dominik Janzing, and Bernhard Sch{{\"o}}lkopf.
\newblock Causal discovery with continuous additive noise models.
\newblock \emph{Journal of Machine Learning Research}, 15\penalty0 (58):\penalty0 2009--2053, 2014.
\newblock URL \url{http://jmlr.org/papers/v15/peters14a.html}.

\bibitem[Petri et~al.(2014)Petri, Expert, Turkheimer, Carhart-Harris, Nutt, Hellyer, and Vaccarino]{petri2014introductionHOmotivation_simplicialFiltrationOnNeruons}
G~Petri, P~Expert, F~Turkheimer, R~Carhart-Harris, D~Nutt, P~J Hellyer, and F~Vaccarino.
\newblock Homological scaffolds of brain functional networks.
\newblock \emph{Journal of The Royal Society Interface}, 2014.

\bibitem[Rolland et~al.(2022)Rolland, Cevher, Kleindessner, Russell, Janzing, Sch{\"o}lkopf, and Locatello]{rolland2022SCORE_leafDetectionAlgorithm}
Paul Rolland, Volkan Cevher, Matth{\"a}us Kleindessner, Chris Russell, Dominik Janzing, Bernhard Sch{\"o}lkopf, and Francesco Locatello.
\newblock Score matching enables causal discovery of nonlinear additive noise models.
\newblock In \emph{Proceedings of the 39th International Conference on Machine Learning}, volume 162 of \emph{Proceedings of Machine Learning Research}, pages 18741--18753. PMLR, 17--23 Jul 2022.

\bibitem[Rosenblatt(1956)]{rosenblatt1956densityFunctionEstimation}
Murray Rosenblatt.
\newblock {Remarks on Some Nonparametric Estimates of a Density Function}.
\newblock \emph{The Annals of Mathematical Statistics}, 27\penalty0 (3):\penalty0 832 -- 837, 1956.
\newblock \doi{10.1214/aoms/1177728190}.
\newblock URL \url{https://doi.org/10.1214/aoms/1177728190}.

\bibitem[Runge et~al.(2019)Runge, Bathiany, Bollt, Camps-Valls, Coumou, Deyle, Glymour, Kretschmer, Mahecha, Mu{\~n}oz-Mar{\'i}, van Nes, Peters, Quax, Reichstein, Scheffer, Scholkopf, Spirtes, Sugihara, Sun, Zhang, and Zscheischler]{runge2019inferring}
Jakob Runge, Sebastian Bathiany, Erik~M. Bollt, Gustau Camps-Valls, Dim Coumou, Ethan~R. Deyle, Clark Glymour, Marlene Kretschmer, Miguel~D. Mahecha, Jordi Mu{\~n}oz-Mar{\'i}, Egbert~H. van Nes, J.~Peters, Rick Quax, Markus Reichstein, Marten Scheffer, Bernhard Scholkopf, Peter Spirtes, George Sugihara, Jie Sun, Kun Zhang, and Jakob Zscheischler.
\newblock Inferring causation from time series in {Earth} system sciences.
\newblock \emph{Nature Communications}, 10, 2019.

\bibitem[Sachs et~al.(2005)Sachs, Perez, Pe'er, Lauffenburger, and Nolan]{sachs2005causal}
Karen Sachs, Omar Perez, Dana Pe'er, Douglas~A Lauffenburger, and Garry~P Nolan.
\newblock Causal protein-signaling networks derived from multiparameter single-cell data.
\newblock \emph{Science}, 308\penalty0 (5721):\penalty0 523--529, 2005.

\bibitem[Schwarz(1978)]{schwarz1978estimating}
Gideon Schwarz.
\newblock Estimating the dimension of a model.
\newblock \emph{The Annals of Statistics}, 6\penalty0 (2):\penalty0 461--464, 1978.

\bibitem[Shimizu et~al.(2006)Shimizu, Hoyer, Hyv\"{a}rinen, and Kerminen]{shimizu2006lingam}
Shohei Shimizu, Patrik~O. Hoyer, Aapo Hyv\"{a}rinen, and Antti Kerminen.
\newblock A linear non-gaussian acyclic model for causal discovery.
\newblock \emph{Journal of Machine Learning Research}, 7\penalty0 (72):\penalty0 2003--2030, 2006.
\newblock URL \url{http://jmlr.org/papers/v7/shimizu06a.html}.

\bibitem[Singh and Moore(2005)]{singh2005finding}
Ajit~P. Singh and Andrew~W. Moore.
\newblock Finding optimal {Bayesian} networks by dynamic programming.
\newblock Technical report, Carnegie Mellon University, 2005.

\bibitem[Spirtes et~al.(2001)Spirtes, Glymour, and Scheines]{spirtes2001causation}
P.~Spirtes, C.~Glymour, and R.~Scheines.
\newblock \emph{Causation, Prediction, and Search}.
\newblock MIT press, 2nd edition, 2001.

\bibitem[Spirtes and Glymour(1991)]{spirtes1991pc}
Peter Spirtes and Clark Glymour.
\newblock An algorithm for fast recovery of sparse causal graphs.
\newblock \emph{Social Science Computer Review}, 9:\penalty0 62--72, 1991.

\bibitem[Taylor and Ehrenreich(2015)]{taylor2015introductionHOmotivation_HOgeneticInteractions}
Matthew~B Taylor and Ian~M Ehrenreich.
\newblock Higher-order genetic interactions and their contribution to complex traits.
\newblock \emph{Trends in Genetics}, 2015.

\bibitem[Tsang et~al.(2020)Tsang, Rambhatla, and Liu]{tsang2020archipelago}
Michael Tsang, Sirisha Rambhatla, and Yan Liu.
\newblock How does this interaction affect me? interpretable attribution for feature interactions.
\newblock \emph{arXiv preprint arXiv:2006.10965}, 2020.

\bibitem[Verma(1993)]{verma1993causalModelTechnicalReport}
T.~S. Verma.
\newblock Graphical aspects of causal models.
\newblock Technical report, UCLA Cognitive Systems Laboratory, 1993.

\bibitem[Verma and Pearl(1990)]{vermaPearl1990skeletonsAndColliders}
Thomas Verma and Judea Pearl.
\newblock Equivalence and synthesis of causal models.
\newblock In \emph{Proceedings of the Sixth Annual Conference on Uncertainty in Artificial Intelligence}, UAI '90, page 255–270, USA, 1990. Elsevier Science Inc.
\newblock ISBN 0444892648.

\bibitem[Xu et~al.(2022)Xu, Marx, Mian, and Vreeken]{xu2022causal}
Sascha Xu, Alexander Marx, Osman Mian, and Jilles Vreeken.
\newblock Causal inference with heteroscedastic noise models.
\newblock In \emph{Proceedings of the AAAI Workshop on Information Theoretic Causal Inference and Discovery}, 2022.

\bibitem[Yuan et~al.(2011)Yuan, Malone, and Wu]{yuan2011learning}
Changhe Yuan, Brandon Malone, and Xiaojian Wu.
\newblock Learning optimal {Bayesian} networks using {A*} search.
\newblock In \emph{International Joint Conference on Artificial Intelligence}, 2011.

\bibitem[Zhang and Hyv\"{a}rinen(2009)]{zhang2009identifiability}
Kun Zhang and Aapo Hyv\"{a}rinen.
\newblock On the identifiability of the post-nonlinear causal model.
\newblock In \emph{Conference on Uncertainty in Artificial Intelligence}, 2009.

\bibitem[Zheng et~al.(2020)Zheng, Dan, Aragam, Ravikumar, and Xing]{zheng2020learning}
Xun Zheng, Chen Dan, Bryon Aragam, Pradeep Ravikumar, and Eric~P. Xing.
\newblock Learning sparse nonparametric {DAGs}.
\newblock In \emph{International Conference on Artificial Intelligence and Statistics}, 2020.

\bibitem[Zheng et~al.(2023)Zheng, Ng, Fan, and Zhang]{zheng2023generalizedPrecisionMatrix}
Yujia Zheng, Ignavier Ng, Yewen Fan, and Kun Zhang.
\newblock Generalized precision matrix for scalable estimation of nonparametric markov networks.
\newblock In \emph{The Eleventh International Conference on Learning Representations}, 2023.
\newblock URL \url{https://openreview.net/forum?id=qBvBycTqVJ}.

\end{thebibliography}
